\documentclass[10pt,journal,compsoc]{IEEEtran}

\ifCLASSOPTIONcompsoc

\else

\fi

\ifCLASSINFOpdf

\else

\fi

\usepackage{epsfig}
\usepackage{url}
\usepackage{verbatim}
\usepackage{graphics}
\usepackage{amsmath}
\usepackage{amssymb}
\usepackage{float}
\usepackage[usenames]{color}
\usepackage{algorithm}
\usepackage{algorithmic}
\usepackage{subfigure}
\usepackage{cite,psfrag}

\usepackage{amsmath}
\usepackage{amssymb}
\usepackage{multirow}
\usepackage{amsmath}
\usepackage{cases}
\usepackage{picins}
\usepackage{float}
\usepackage{caption}
\usepackage{graphics}
\begin{document}
\graphicspath{{figures/}}
\newcommand{\eqs}{\small}
\newcommand{\sps}{\scriptsize}%
\newcommand{\Tabsize}{\footnotesize}
\newcommand{\Captionsize}{\small}

\title{Infinite Mixture of Inverted Dirichlet Distributions}
\author{Zhanyu~Ma and Yuping~Lai
\IEEEcompsocitemizethanks{
\IEEEcompsocthanksitem Preliminary work of ongoing work.}
\thanks{}}
%

\markboth{}%
{Shell \MakeLowercase{\textit{et al.}}: Bare Advanced Demo of IEEEtran.cls for Journals}

\IEEEcompsoctitleabstractindextext{%
\begin{abstract}
In this work, we develop a novel Bayesian estimation method for the Dirichlet process (DP) mixture of the inverted Dirichlet distributions, which has been shown to be very flexible for modeling vectors with positive elements. The recently proposed extended variational inference (EVI) framework is adopted to derive an analytically tractable solution. The convergency of the proposed algorithm is theoretically guaranteed by introducing single lower bound approximation to the original objective function in the VI framework. In principle, the proposed model can be viewed as an infinite inverted Dirichelt mixture model (InIDMM) that allows the automatic determination of the number of mixture components from data. Therefore, the problem of pre-determining the optimal number of mixing components has been overcome. Moreover, the problems of over-fitting and under-fitting are avoided by the Bayesian estimation approach. Comparing with several recently proposed DP-related methods, the good performance and effectiveness of the proposed method have been demonstrated with both synthesized data and real data evaluations.
\end{abstract}

}

\maketitle

\IEEEdisplaynotcompsoctitleabstractindextext

\IEEEpeerreviewmaketitle

\section{Introduction}
\IEEEPARstart{F}inite mixture modeling~\cite{01,02,28} is a flexible and powerful probabilistic modeling tool for univariate and multivariate data that are assumed to be generated from heterogeneous populations. It has been widely applied to many areas, such as pattern recognition, machine learning, data mining, computer vision~\cite{03}. Among all finite mixture models, the finite Gaussian mixture model (GMM) has been the most popular method for modeling continuous data. Much of its popularity is due to the fact that any continuous distribution can be arbitrarily well approximated by a GMM with unlimited number of mixture components. Moreover, the parameters in a GMM can be estimated efficiently via maximum likelihood (ML) estimation with the expectation maximum (EM) algorithm~\cite{Reynolds1995}. By assigning prior distributions to the parameters in a GMM, Bayesian estimation of GMM can be carried out with conjugate prior-posterior pair matching~\cite{Nasios2006}. Both the ML and the Bayesian estimation algorithms can be represented by an analytically tractable form~\cite{Nasios2006}.

Recent studies have shown that non-Gaussian statistical models,~\emph{e.g.}, the beta mixture model (BMM), the Dirichlet mixture model (DMM), the Gamma mixture model (GaMM)~\cite{Jung2014}, the von Mises-Fisher mixture model (vMM)~\cite{Taghia2014}, can model the non-Gaussian distributed data more efficiently, compared to the conventional GMM. For example, BMM has been widely applied in modeling grey image pixel values and DNA methylation data~\cite{Houseman2008}. In order to efficiently model proportional data~\cite{Nascimento2012,Ma2013}, DMM can be utilized to describe the underlying distribution. In generalized-$K$ ($K_G$) fading channels, GaMM has been used to analyze the capacity and error probability~\cite{Jung2014}. The vMM has been widely used in modeling directional data, such as yeast gene expression~\cite{Taghia2014} and topic detection~\cite{He2010}. The finite inverted Dirichlet mixture model (IDMM), among others, has been demonstrated to be an efficient tool for modeling data vector with positive elements~\cite{10,Bdiri2013}. For instance, IDMM has been widely used for software module classification~\cite{10}, visual scene analysis and classification~\cite{Bdiri2013a,Bdiri2013}.

An essential problem in finite mixture modeling, no matter with Gaussian or non-Gaussian mixture component, is how to automatically decide the appropriate number of mixture components based on the data. The component number has a strong effect on the modeling accuracy~\cite{Markley2010}. If the number of mixture components is not properly chosen, the mixture model may over-fit or under-fit the observed data. To deal with this problem, many methods have been proposed. These can be categorized into two groups: deterministic approaches~\cite{11,12} and Bayesian methods~\cite{Richardson1996}. Deterministic approaches are generally implemented by ML estimation under an EM-based and require the integration of entropy measures or some information theoretic criteria, such as the minimum message length (MML)~\cite{12}, the Bayesian information criterion (BIC)~\cite{Huang2016}, and the Akaike information criterion (AIC)~\cite{Chen2013}, to determine the number of components in the mixture model. It is worth noting that, in general, the EM algorithm converges to a local maximum or a saddle point and its solution is highly dependent on its initialization. On the other hand, the Bayesian methods, which are not sensitive to initialization by introducing proper prior distributions to the parameters in the model, have been widely used to find a suitable number of components in a finite mixture model. In this case, the parameters of a finite mixture model (including the parameters in a component and the weighting coefficients) are treated as random variables under the Bayesian framework. The posterior distributions of the parameters, rather than simple point estimates, are computed~\cite{28}. The model truncation in Bayesian estimation of finite mixture model is carried out by setting the corresponding weights of the unimportant mixture components to zero (or a small value close to zero)~\cite{28}. However, the number of mixture components should be properly initialized, as it can only decrease during the training process.

The increasing interest in mixture modeling has led to the development of the model selection method\footnotemark\footnotetext{Here, model selection means selecting the best of a set of models of different orders}. Recent work has shown that the non-parametric Bayesian approach~\cite{Bousmalis2015,Meila2016} can provide an elegant solution for automatically determining the complexity of model. The basic idea behind this approach is that it provides methods to adaptively select the optimal number of mixing components, while also allows the number of mixture components to remain unbounded. In other words, this approach allows the number of components to increase as new data arrives, which is the key difference from finite mixture modeling. The most widely used Bayesian nonparametric~\cite{Hjort2010} model selection method is based on the Dirichlet process (DP) mixture model~\cite{20,Foti2015}. The DP mixture model extends distributions over measures, which has the appealing property that it does not need to set a prior on the number of components. In essence, the DP mixture model can also be viewed as an infinite mixture model with its complexity increasing as the size of dataset grows. Recently, the DP mixture model has been applied in many important applications. For instance, the DP mixture model has been adopted to a mixture of different types of non-Gaussian distributions, such as the DP mixture of beta-Liouville distributions~\cite{17}, the DP mixture of student's-t distributions~\cite{21}, the DP mixture of generalized Dirichlet distributions~\cite{22}, the DP mixture of student's-t factors~\cite{23}, and the DP mixture of hidden Markov random field models~\cite{24}.

Generally speaking, most parameter estimation algorithms for both the deterministic and the Bayesian methods are time consuming, because they have to numerically evaluate a given model selection criterion~\cite{Bouguila2007,Bouguila2012}. This is especially true for the fully Bayesian Markov chain Monte Carlo (MCMC)~\cite{14,Meila2016}, which is one of the widely applied Bayesian approaches with numerical simulations. The MCMC approach has its own limitations, when high-dimensional data are involved in the training stage~\cite{Robert2007,Pereyra2016}. This is due to the fact that its sampling-based characteristics yield a heavy computational burden and it is difficult to monitor the convergence in the high-dimensional space. To overcome the aforementioned problems, variational inference (VI), which can provide an analytically tractable solution and good generalization performance, has been proposed as an efficient alternative to the MCMC approach~\cite{15,Ma2010a}. With an analytically tractable solution, the numerical sampling during each iteration in the optimization stage can be avoided. Hence, the VI-based solutions can lead to more efficient estimation. They have been successfully applied in a variety of applications including the estimation of mixture models~\cite{06}.

Motivated by the ability of the Bayesian non-parametric approaches to solve the model selection problem and the good performance recently obtained by the VI framework, we focus on the variational learning of the DP mixture of inverted Dirichlet distributions (\emph{a.k.a.} the infinite inverted Dirichlet mixture model (InIDMM)). Since InIDMM is a typical non-Gaussian statistical model, it is not feasible to apply the traditional VI framework to obtain an analytically tractable solution for the Bayesian estimation. To derive an analytically tractable solution for the variational learning of InIDMM, the recently proposed extended variational inference (EVI), which is particularly suitable for non-Gaussian statistical models, has been adopted to provide an appropriate~\emph{single lower bound approximation} to the original object function. With the auxiliary functiond, an analytically tractable solution for Bayesian estimation of InIDMM is derived. The key contributions of our work are three-fold: 1) The finite inverted Dirichlet mixture model (IDMM) has been extended to the infinite inverted Dirichlet mixture model (InIDMM) under the stick-breaking framework~\cite{20,Paisley2015}. Thus, the difficulty in automatically determining the number of mixture components can be overcome. 2) An analytically solution is derived with the EVI framework for InIDMM. Moreover, comparing with the recently proposed algorithm for InIDMM~\cite{Fan2015}, which is based on~\emph{multiple lower bound approximation}, our algorithm can not only theoretically guarantee convergence but also provide better approximations. 3) The proposed method has been applied in several important applications, such as image categorization and object detection. The good performance has been illustrated with both synthesized and real data evaluations.

The remaining part of this paper is organized as follow: Section~\ref{Sec: Statistical Model} provides a brief overview of the finite inverted Dirichlet mixture and the DP mixture. The infinite inverted Dirichlet mixture model is also proposed. In Section~\ref{Sec: VI for InIDMM}, a Bayesian learning algorithm with EVI is derived. The proposed algorithm has an analytically tractable form. The experimental results with both synthesized and real data evaluations are reported in Section~\ref{Sec: Experiments}. Finally, we draw conclusions and future research directions in Section~\ref{Sec: Conclusion}.

\section{The statistical model}
\label{Sec: Statistical Model}
In this section, we first present a brief overview of the finite inverted Dirichlet mixture model (IDMM). Then, the DP mixture model with stick-breaking representation is introduced. Finally, we extend the IDMM to InIDMM.
\subsection{Finite inverted Dirichlet mixture model}
Given a \emph{D}-dimensional vector ${\vec x} = \{ {x_{1}}, \cdots ,{x_{D}}\}$ generated from a IDMM with ${M}$ components, the  probability density function (PDF) of $\vec x$ is denoted as~\cite{10}
\begin{equation}\eqs\label{eq:1}
\text{IDMM}(\vec{x}|\vec{\pi },{\Lambda })=\sum\limits_{m=1}^{M}{{{\pi }_{m}}}\text{iDir}(\vec{x}|{{\vec{\alpha }}_{m}}),
\end{equation}
where ${\Lambda}=\{{{\vec{\alpha}}_{m}}\}_{m=1}^{M}$ and $\vec \pi = \{\pi_m\}_{m=1}^{M}$ is the mixing coefficient vector subject to the constraints $0\le {{\pi }_{m}}\le 1$ and $\sum\nolimits_{m=1}^{M}{{{\pi }_{m}}=1}$. Moreover, $\text{iDir}(\vec{x}|\vec{\alpha})$ is an inverted Dirichlet distribution with its $(D+1)$-dimensional positive parameter vector $\vec{\alpha }=\{{{\alpha }_{1}},\cdots ,{{\alpha }_{D+1}}\}$ defined as
\begin{equation}\eqs\label{eq:2}
\text{iDir}(\vec{x}|\vec{\alpha })=\frac{\Gamma (\sum\nolimits_{d=1}^{D+1}{{{\alpha }_{d}}})}{\prod\nolimits_{d=1}^{D+1}{\Gamma ({{\alpha }_{d}})}}\prod\limits_{d=1}^{D}{x_{d}^{{{\alpha }_{d}}-1}{{\left( 1+\sum\limits_{d=1}^{D}{{{x}_{d}}} \right)}^{-\sum\nolimits_{d=1}^{D+1}{{{\alpha }_{d}}}}}},
\end{equation}
where ${{x}_{d}}>0$ for $d=1,\cdots,D$ and $\Gamma (\cdot)$ is the Gamma function defined as $\Gamma (a)=\int_{0}^{\infty }{{{t}^{a-1}}}{{e}^{-t}}dt$.

\subsection{Dirichlet Process with Stick-Breaking}
\label{Sec: DP with SB}
The Dirichlet process (DP)~\cite {20,Foti2015} is a stochastic process used for Bayesian nonparametric data analysis, particularly in a DP mixture model (infinite mixture model). It is a distribution over distributions rather than parameters,~\emph{i.e.}, each draw from a DP is a probability distribution itself, rather than a parameter vector~\cite{Frigyik2010}. We adopt the DP to extend the IDMM to the infinite case, such that the difficulty of the automatic determination of the model complexity (\emph{i.e.}, the number of mixture components) can be overcome. To this end, the DP is constructed by the following stick-breaking formulation~\cite{Hjort2010}, which is an intuitive and simple constructive definition of the DP.

Assume that ${H}$ is a random distribution and $\varphi$ is a positive real scalar. We consider two countably infinite collections of independently generated stochastic variables $ {{\Omega }_{m}}\sim H$ and ${{\lambda }_{m}}\sim \text{Beta}({\lambda }_{m};1,\varphi )$\footnotemark\footnotetext{To avoid confusion, we use $f\left(x;a\right)$ to denote the PDF of $x$ parameterized by parameter $a$. $f\left(x|a\right)$ is used to denote the conditional PDF of $x$ given $a$, where both $x$ and $a$ are random variables. Both $f\left(x;a\right)$ and $f\left(x|a\right)$ have exactly the same mathematical expressions.} for $m=\{1,\cdots,\infty \}$, where $\text{Beta}(x;a,b)$ is the beta distribution defined as $\text{Beta}(x;a,b)=\frac{\Gamma(a+b)}{\Gamma(a)\Gamma(b)}x^{a-1}(1-x)^{b-1}$. A distribution $G$ is said to be DP distributed with a concentration parameter $\varphi$ and a base measure or base distribution \emph{H} (denoted as $G\sim \text{DP}(\varphi ,H))$, if the following conditions are satisfied:
\begin{equation}\eqs\label{eq:3}
G=\sum\limits_{m=1}^{\infty }{{{\pi }_{m}}{{\delta }_{{{\Omega }_{m}}}}},\ {{\pi }_{m}}={{\lambda }_{m}}\prod\limits_{l=1}^{m-1}{(1-{{\lambda }_{l}})},
\end{equation}
where $\{{{\pi }_{m}}\}$ is a set of stick-breaking weights with constraints $\sum\nolimits_{m=1}^{\infty }{{{\pi }_{m}}=1}$, ${{\delta }_{{{\Omega }_{m}}}}$ is a delta function whose value is $1$ at location ${{\Omega }_{m}}$ and $0$ otherwise. The generation of the mixing coefficients $\{{{\pi }_{m}}\}$ can be considered as process of breaking a unit length stick into an infinite number of pieces. The length of each piece, $\lambda_m$, which is proportional to the rest of the ``stick" before the current breaking, is considered as an independent random variable generated from $\text{Beta}(\lambda_m;1,\varphi )$. Because of its simplicity and natural generalization ability, the stick-breaking construction has been a widely applied scheme for the inference of DPs~\cite{17,Paisley2015,Paisley2009}.

\subsection{Infinite Inverted Dirichlet Mixture Model}
Now we consider the problem of modeling $\vec x$ by an Infinite Inverted Dirichlet Mixture Model (InIDMM), which is actually an extended IDMM with an infinite number of components. Therefore,~\eqref{eq:1} can be reformulated as
\begin{equation}\eqs
\label{eq:4}
\text{InIDMM}(\vec{x}|\vec{\pi },\Lambda )=\sum\limits_{m=1}^{\infty }{{{\pi }_{m}}}\text{iDir}(\vec{x}|{{\vec{\alpha }}_{m}}),
\end{equation}
where $\vec{\pi}=\{\pi_m\}_{m=1}^{\infty}$ and $\Lambda=\{\vec{\alpha}_m\}_{m=1}^{\infty}$
Then, the likelihood function of the InIDMM given the observed dataset $\mathcal X=\{{{\vec{x}}_{n}}\}_{n=1}^{N}$ is given by
\begin{equation}\eqs\label{eq:5}
\text{InIDMM}(\mathcal X|\vec{\pi },\Lambda )=\prod\limits_{n=1}^{N}{\left\{ \sum\limits_{m=1}^{\infty }{{{\pi }_{m}}}\text{iDir}({{{\vec{x}}}_{n}}|{{{\vec{\alpha }}}_{m}}) \right\}}.
\end{equation}
In order to clearly illustrate the generation process of each observation ${{\vec{x}}_{n}}$ in the mixture model, we introduce a latent indication vector variable
$\vec z_n=\{z_{n1},z_{n2}, \cdots\}$. $\vec z$ has only one element equal to $1$ and the other elements in $\vec z$ are $0$. For example, ${{{z}}_{nm}}=1$ indicates the sample ${{\vec{x}}_{n}}$ comes from the mixture component $m$. Therefore, the conditional distribution of $\mathcal X$ given the parameters $\Lambda$ and the latent variables $\mathcal{Z}=\{{{{z}}_{nm}}\}$ is
\begin{equation}\eqs\label{eq:6}
\text{InIDMM}(\mathcal X|\mathcal Z,\Lambda )=\prod\limits_{n=1}^{N}{\prod\limits_{m=1}^{\infty }{\text{iDir}{{({{{\vec{x}}}_{n}}|{{\vec\alpha }_{m}})}^{{{z}_{nm}}}}}}.
\end{equation}
Moreover, to exploit the advantages of the Bayesian framework, conjugate prior distributions are introduced for all the unknown parameters according to their distribution properties. In this work, we place the conjugate priors over the unknown stochastic variables $\mathcal Z$, $\Lambda$, and $\vec{\lambda }=({{\lambda }_{1}}, {{\lambda }_{2}},\cdots )$ such that a full Bayesian estimation model can be obtained.

In the aforementioned full Bayesian model, the prior distribution of $\mathcal Z$ given $\vec\pi$ is given by
\begin{equation}\eqs\label{eq:7}
p(\mathcal Z|\vec{\pi })=\prod\limits_{n=1}^{N}{\prod\limits_{m=1}^{\infty }{\pi _{m}^{{{{z}}_{nm}}}}}.
\end{equation}
As $\vec\pi$ is a function of $\vec{\lambda }$ according to the stick-breaking construction of the DP as shown in~\eqref{eq:3}, we rewrite~\eqref{eq:7} as
\begin{equation}\eqs\label{eq:8}
p(\mathcal{Z}|\vec{\lambda })={{\prod\limits_{n=1}^{N}{\prod\limits_{m=1}^{\infty }{\left[ {{\lambda }_{m}}\prod\limits_{l=1}^{m-1}{(1-{{\lambda }_{l}})} \right]}}}^{{{{z}}_{nm}}}}.
\end{equation}

As previously mentioned in Section~\ref{Sec: DP with SB}, the prior distribution of $\vec\lambda$ is
\begin{equation}\eqs\label{eq:9}
p(\vec{\lambda }|\vec{\varphi })=\prod\limits_{m=1}^{\infty }{\text{Beta}(\lambda_m;1,{{\varphi }_{m}})}=\prod\limits_{m=1}^{\infty }{{{\varphi }_{m}}}{{(1-{{\lambda }_{m}})}^{{{\varphi }_{m}}-1}},
\end{equation}
where $\vec{\varphi }=({{\varphi}_{1}},{{\varphi }_{2}},\cdots )$. Based on~\eqref{eq:3}, we can obtain the expected value of $\pi_{m}$. In order to do this, the expected value of ${{\lambda }_{m}}$ will first be calculated as
\begin{equation}\eqs\label{eq: Elambda}
\langle {{\lambda }_{m}}\rangle =1/(1+{{\varphi }_{m}}).
\end{equation}
Then, the expected value of $\pi_{m}$ is denoted as
\begin{equation}\eqs\label{eq: pi}
\langle {{\pi}_{m}}\rangle =\langle {{\lambda}_{m}}\rangle \prod_{l=1}^{m-1}(1-\langle{\lambda}_{l}\rangle).
\end{equation}
It is worth to note that, when the value of $\varphi_{m}$ is small, $\langle {{\lambda }_{m}}\rangle $ will become large. Therefore, the expected of the mixing coefficients ${\pi_{m}}$ are controlled by the parameters $\varphi_{m}$,~\emph{i.e.}, small value of $\varphi_{m}$ will yield small ${\pi_{m}}$ such that the distribution of ${\pi_{m}}$ will be sparse.

As $\varphi_m$ is positive, we assume $\vec \varphi$ follows a product of gamma prior distributions as
\begin{equation}\eqs\label{eq:10}
p(\vec{\varphi };\vec s,\vec t)=\prod\limits_{m=1}^{\infty }{\text {Gam}}({\varphi_m};{s_m},{t_m})=\prod\limits_{m=1}^{\infty }{\frac{t_{m}^{{{s}_{m}}}}{\Gamma ({{s}_{m}})}}\varphi _{m}^{{{s}_{m}}-1}{{e}^{-{{t}_{m}}{{\varphi }_{m}}}},
\end{equation}
where ${\text {Gam}}(\cdot )$ is the gamma distribution. $\vec{s}=({{s}_{1}},{{s}_{2}},\cdots )$ and $\vec{t}=({{t}_{1}},{{t}_{2}},\cdots )$ are the hyperparamters and subject to the constraints ${{s}_{m}}>0$ and ${{t}_{m}}>0$.

Next, we introduce an approximating conjugate prior distribution to parameter $\Lambda$ in InIDMM. The inverted Dirichlet distribution belongs to the exponential family and its formal conjugate prior can be derived with the Bayesian rule~\cite{28} as
\begin{equation}\eqs\label{eq:47}
p(\vec{\alpha }|{{\vec{\mu }}_{0}},{{v}_{0}})=C({{\vec{\mu }}_{0}},{{v}_{0}}){{\left[ \frac{\Gamma (\sum\nolimits_{d=1}^{D+1}{{{\alpha }_{d}}})}{\prod\nolimits_{d=1}^{D+1}{{{\alpha }_{d}}}} \right]}^{{{\nu }_{0}}}}{{e}^{-{{{\vec{\mu }}}_{0}}({{{\vec{\alpha }}}^{T}}-\vec{I}_{D+1})}},
\end{equation}
where ${{\vec{\mu }}_{0}}=[{{\mu }_{{{1}_{0}}}},\cdots {{\mu }_{D+{{1}_{0}}}}]$ and ${{\nu }_{0}}$ are the hyperparameters in the prior distribution, $ C({{\vec{\mu }}_{0}},{{v}_{0}})$ is a normalization coefficient such that $\int{p(\vec{\alpha }|{{{\vec{\mu }}}_{0}},{{v}_{0}})d\vec{\alpha }}=1$. ${{\vec {I}}_{d}}$ is a $D$-dimensional vector with all elements equal to one. Then, we can write the posterior distribution of $\vec\alpha$ as (with \emph{N i.i.d.} observations  $\mathcal X$)
\begin{equation}\eqs\label{eq:48}
\begin{split}
& f(\vec{\alpha }|\mathcal X)=\frac{\text{iDir}(\mathcal X|\vec{\alpha })f(\vec\alpha|{{{\vec{\mu }}}_{0}},{{\nu }_{0}})}{\int{\text{iDir}(\mathcal X|\vec{\alpha })f(\vec\alpha|{{{\vec{\mu }}}_{0}},{{\nu }_{0}})d\vec{\alpha }}}\  \\
& \ \ \ \ \ \ \ \ \ \ \ \ =C({{\vec{\mu }}_{N}},{{\nu }_{N}}){{\left[ \frac{\Gamma (\sum\nolimits_{d=1}^{D+1}{{{\alpha }_{d}}})}{\prod\nolimits_{d=1}^{D+1}{\Gamma ({{\alpha }_{d}})}} \right]}^{{{\nu }_{N}}}}{{e}^{-{{{\vec{\mu }}}_{N}}({{{\vec{\alpha }}}^{T}}-\vec{I}_{D+1})}}
\end{split}
\end{equation}
where the hyperparameters ${{\nu }_{N}}$ and ${{\vec{\mu }}_{N}}$ in the posterior distribution are
\begin{equation}\eqs\label{eq:49}
{{\nu }_{N}}={{\nu }_{0}}+N, {{\vec{\mu }}_{N}}={{\vec{\mu }}_{0}}-{{[\ln \mathcal {X}^+-\vec{I}_{D+1}\ln (1+{{\vec{I}}_{D+1}}^T\mathcal {X}^+)]}}{{\vec{I}}_{N}}.
\end{equation}
In~\eqref{eq:49}, $\mathcal {X}^+$ is a $(D+1)\times N$ matrix by connecting $\vec{I}_{D+1}^T$ to the bottom of $\mathcal {X}$.
However, it is not applicable in our VI framework due to the analytically intractable normalization factor in~\eqref{eq:48}. Because $\Lambda $ is positive, we adopt gamma prior distributions to approximate conjugate prior for $\Lambda$ as well. By assuming the parameters of inverted Dirichlet distribution are mutually independent, we have
\begin{equation}\eqs\label{eq:11}
p(\Lambda )=\text{Gam}(\Lambda ;U,V)=\prod\limits_{m=1}^{\infty }{\prod\limits_{d=1}^{D+1}{\frac{v_{md}^{{{u}_{md}}}}{\Gamma ({{u}_{md}})}}}\alpha _{md}^{{{u}_{md}}-1}{{e}^{-{{v}_{md}}{{\alpha }_{md}}}},
\end{equation}
where all the hyperparameters $U =\{{{u}_{md}}\}$ and $V =\{{{v}_{md}}\}$ are positive.

\begin{figure}[!t]
\vspace{-5mm}
\psfrag{a}[c][c]{ $\vec{z}_n$}
\psfrag{e}[c][c]{ $\infty$}
\psfrag{b}[c][c]{ $\lambda_m$}
\psfrag{c}[c][c]{ $\varphi_m$}
\psfrag{d}[c][c]{ $\vec{\alpha}_m$}
\psfrag{n}[c][c]{ $N$}
\psfrag{x}[c][c]{ $\vec{x}_n$}
\centering\includegraphics[width=.45\textwidth]{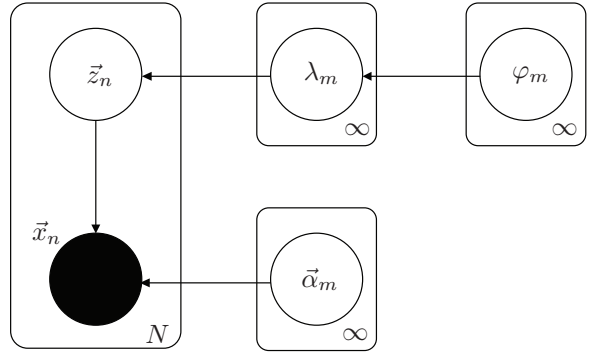}
\caption{\label{Fig: Graphical Model}Graphical representation of the variables¡¯ relationships in the Bayesian inference of a InIDMM.  All of the circles in the graphical figure represent variables.  Arrows show the relationships between variables. The variables in the box are the~\emph{i.i.d.} observations.}
\vspace{-5mm}
\end{figure}

With the Bayesian rules and by combining \eqref{eq:6} and \eqref{eq:8}-\eqref{eq:11} together, we can represent the joint density of the observation $\mathcal X$ with all the $i.i.d.$ latent variables $\Theta = ({\mathcal{Z},\Lambda,\vec{\lambda },\vec{\varphi }})$ as
\begin{equation}\eqs\label{eq:12}
\begin{split}
p(\mathcal{X},\Theta )=&p(\mathcal{X}|\mathcal{Z},\Lambda)p(\mathcal{Z}|\vec{\lambda })p(\vec{\lambda }|\vec{\varphi })p(\vec{\varphi })p(\Lambda)\  \\
 =&\prod\limits_{n=1}^{N}{\prod\limits_{m=1}^{\infty }{\left[ {{\lambda }_{m}}\prod\limits_{j=1}^{m-1}{(1-{{\lambda }_{j}})}\frac{\Gamma \left( \sum\nolimits_{d=1}^{D+1}{{{\alpha }_{md}}} \right)}{\prod\nolimits_{d=1}^{D+1}{\Gamma ({{\alpha }_{md}})}} \right.}} \\
& {{\left.  \times \prod\limits_{d=1}^{D}{x_{nd}^{{{\alpha }_{md}}-1}{{\left( 1+\sum\limits_{d=1}^{D}{{{x}_{nd}}} \right)}^{-\sum\nolimits_{d=1}^{D+1}{{{\alpha }_{md}}}}}} \right]}^{{{z}_{nm}}}} \\
&\times \prod\limits_{m=1}^{\infty }{\left[ {{\varphi }_{m}}{{(1-{{\lambda }_{m}})}^{{{\varphi }_{m}}-1}}\frac{t_{m}^{{{s}_{m}}}}{\Gamma ({{s}_{m}})}\varphi _{m}^{{{s}_{m}}-1}{{e}^{-{{t}_{m}}{{\varphi }_{m}}}} \right]}\  \\
&  \times \prod\limits_{m=1}^{\infty }{\prod\limits_{d=1}^{D+1}{\frac{v_{md}^{{{u}_{md}}}}{\Gamma ({{u}_{md}})}}}\alpha _{md}^{{{u}_{md}}-1}{{e}^{-{{v }_{md}}{{\alpha }_{md}}}}.
\end{split}
\end{equation}
The structure of the InIDMM can be represented in terms of a directed probabilistic graphical model, as shown in Fig.~\ref{Fig: Graphical Model}, which illustrates the relations among the variables and the observations.

\section{Variational Learning for InIDMM}
\label{Sec: VI for InIDMM}
In this section, we develop a variational Bayesian inference framework for learning the InIDMM. With the assistance of recently proposed EVI, an analytically tractable algorithm, which prevents numerical sampling during each iteration and facilitates a training procedure, is obtained. The proposed solution is also able to overcome the problem of overfitting and automatically decide the number of mixture components.
\subsection{Variational Inference}
The purpose of Bayesian analysis is to estimate the values of the hyperparameters as well as the posterior probability distribution of the latent variables. Although we can formulate the posterior distribution $p(\Theta|\mathcal X)$ by using the Bayesian rule as
\begin{equation}\eqs\label{eq:13}
p({\Theta} |\mathcal{X})=\frac{p(\mathcal{X},{\Theta} )}{p(\mathcal{X})}.
\end{equation}
The calculation of $p(\mathcal X)$ from the joint distribution $p(\mathcal X, \Theta)$, which involves the summation and the integration over the latent variables, is analytically intractable for most of non-Gaussian statistical models. Therefore, we apply the VI framework to approximate the actual posterior $p(\mathcal X|\Theta)$ with an approximating distribution $q(\Theta)$. In principle, $q(\Theta)$ can be of arbitrary form. To make this approximation as close as possible to the actual posterior distribution, we can find the optimal approximation by minimizing the Kullback-Leibler (KL) divergence of $p(\Theta|\mathcal X)$ from $q( \Theta)$ as
\begin{equation}\eqs\label{eq:14}
\begin{split}
\text{KL}(q||p)=-\int{q({\Theta} )}\ln \left\{ \frac{p( {\Theta} |\mathcal{X})}{q({\Theta} )}\right\} d\Theta.
\end{split}
\end{equation}
By some mathematical manipulations, we obtain the following expression:
\begin{equation}\eqs\label{eq:15}
\ln p({\mathcal X})=\underbrace{\int{q( \Theta )}\ln \left\{ \frac{p(\mathcal X, \Theta)}{q( \Theta )} \right\} d\Theta}_{\mathcal L(q)}-\underbrace{\int{q( \Theta )}\ln \left\{ \frac{p(\Theta |\mathcal{X})}{q( \Theta )}\right\}d\Theta}_{\text{KL}(q||p)},
\end{equation}
where the KL divergence $\text{KL}(q||p)$ is nonnegative and can be equal to zero~\emph{if and only if} $q(\Theta)=p(\Theta|\mathcal X)$~\cite{27}. However, it is infeasible to solve $q(\Theta)$ by minimizing $\text{KL}(q( {\Theta} )||p({\Theta} |\mathcal{X}))$, as $p(\Theta|\mathcal X)$ is unknown. As the logarithm of the marginal evidence $\ln p({\mathcal X})$ is fixed by a given ${\mathcal X}$, minimizing the KL divergence is equivalent to maximizing the lower bound $\mathcal L(q)$ (which is known as the variational objective function). Hence, it is usual to find the optimal approximation $q( \Theta )$ by maximizing $\mathcal L(q)$~\cite{28} in the VI framework. Within the variational inference framework, the variational objective function that needs to be maximized is
\begin{equation}\eqs\label{eq:16}
\mathcal L(q)={{\text{E}}_{q(\Theta)}}[\ln p(\mathcal X,\Theta)]-{{\text{E}}_{q(\Theta)}}[\ln q(\Theta)].
\end{equation}
\subsection{Extended Variational Inference}
For most of the non-Gaussian mixture models (\emph{e.g.}, the beta mixture model, the Dirichlet mixture model, the beta-Liouville mixture model~\cite{17}, the inverted Dirichlet mixture model~\cite{Bdiri2013}), the term ${\text{E}_{q( \Theta)}}[\ln p(\mathcal X, \Theta)]$ is analytically intractable such that the lower bound $\mathcal L(q)$ cannot be maximized directly by a closed-form solution. Therefore, the EVI method \cite{15,16} was proposed to overcome the aforementioned problem. With an auxiliary function $\tilde{p}(\mathcal X, \Theta )$ that satisfies
\begin{equation}\eqs\label{eq:18}
{{\text{E}}_{q( \Theta )}}[\ln p(\mathcal{X}, \Theta )]\ge {{\text{E}}_{q( \Theta )}}[\ln \tilde{p}(\mathcal{X}, \Theta )]
\end{equation}
and substituting~\eqref{eq:18} into~\eqref{eq:16}, we can still reach the maximum value of $\mathcal L(q)$ at some given points by maximizing a lower bound of $\tilde{\mathcal L}(q)$
\begin{equation}\eqs\label{eq:17}
\mathcal L(q)\ge \tilde{\mathcal L}(q)={{\text{E}}_{q(\Theta )}}[\ln \tilde{p}(\mathcal{X}, \Theta )]-{{\text{E}}_{q( \Theta )}}[\ln q( \Theta )].
\end{equation}
If $\tilde{p}(\mathcal X, \Theta )$ is properly selected, an analytically tractable solution can be obtained. The strategy for selecting a proper $\tilde{p}(\mathcal X, \Theta )$ can be found.

In order to properly formulate the variational posterior $q(\Theta)$, we truncate the stick-breaking representation for the InIDMM at a value ${M}$ as
\begin{equation}\eqs\label{eq:19}
{\lambda _M} = 1,\;\;\;\;{\pi _m} = 0 \;\;\;\; \text {when} \; m > M, \;\;\;\; \text{and} \sum\limits_{m = 1}^M {{\pi _m}}  = 1.
\end{equation}
Note that the model is still a full DP mixture. The truncation level ${M}$ is not a part of our prior infinite mixture model, it is only a variational parameter for pursuing an approximation to the posterior, which can be freely initialized and automatically optimized without yielding overfitting during the learning process. Additionally, we make use of the following factorized variational distribution to approximate $p( \Theta |\mathcal{X})$ as
\begin{equation}\eqs\label{eq:20}
\begin{split}
q(\Theta ) =&\left[ \prod\limits_{m=1}^{M}{q({{\lambda }_{m}})q({{\varphi }_{m}})} \right]\left[ \prod\limits_{n=1}^{N}{\prod\limits_{m=1}^{M}{q({{{z}}_{nm}})}} \right] \\
    & \times \left[ \prod\limits_{m=1}^{M}{\prod\limits_{d=1}^{D+1}{q({{\alpha}_{md}})}} \right],
\end{split}
\end{equation}
where the variables in the posterior distribution are assumed to be mutually independent (as illustrated by the graphical model in Fig.~\ref{Fig: Graphical Model}). This is the only assumption we introduced to the posterior distribution. No other restrictions are imposed over the mathematical forms of the individual factor distributions~\cite{28}.

Applying the full factorization formulation and the truncated stick-breaking representation for the proposed model, we can solve the variational learning by maximizing the lower bound  $\tilde{\mathcal L}(q)$ shown in \eqref{eq:17}. The optimal solution in this case is given by
\begin{equation}\eqs\label{eq:21}
\ln {{q}_{s}}({{\Theta }_{s}})={{\left\langle \ln \tilde{p}(\mathcal{X}, \Theta ) \right\rangle }_{j\ne s}}+\text{Con.},
\end{equation}
where ${{\langle \cdot \rangle }_{j\ne s}}$ refers to the expectation with respect to all the distributions ${q_j}({\Theta _j})$ except for variable $s$. In addition, any term that does not include $\Theta_{s}$ are absorbed into the additive constant ``Con.''~\cite{15,28}. In the variational inference, all factors ${q_s}({\Theta _s})$ need to be suitably initiated, then each factor is updated in turn with a revised value obtained by~\eqref{eq:21} using the current values of all the other factors. Convergence is theoretically guaranteed since the lower bound is a convex with respect to each factor ${{q}_{s}}({{\Theta }_{s}})$~\cite{28}. It is is worth noting that, although convergence is promised, the algorithm may also fall in local maxima or saddle points.

\subsection{EVI for the Optimal Posterior Distributions}
According to the principles of EVI, the expectation of the logarithm of the joint distribution, given the joint posterior distributions of the parameters, can be expressed as
\begin{equation}\eqs\label{eq:22-0}
\begin{split}
& \langle \ln {p}(\mathcal{X},\Theta )\rangle\\
 =&\sum\limits_{n=1}^{N}{\sum\limits_{m=1}^{M}{\langle {{z}_{nm}}\rangle }}\left[ {{{{\mathcal{R}}}}_{m}}\text{+}\sum\limits_{d=1}^{D}{(\langle{{{\alpha }}}_{md}\rangle-1)\ln {{x}_{nd}}} \right. \\
& \left. -\sum\limits_{d=1}^{D+1}{\langle{{{{\alpha }}}_{md}}\rangle}(1+\sum\limits_{d=1}^{D}{{{x}_{nd}}})+\langle \ln {{\lambda }_{m}}\rangle +\sum\limits_{j=1}^{m-1}{\langle \ln (1-{{\lambda }_{j}})\rangle }\right] \\
&+\sum\limits_{m=1}^{M}{\left[ \langle \ln {{\varphi }_{m}}\rangle +(\langle {{\varphi }_{m}}\rangle -1)\langle \ln (1-{{\lambda }_{m}})\rangle  \right]}\  \\
&+\sum\limits_{m=1}^{M}{\sum\limits_{d=1}^{D+1}{\left[ ({{u}_{md}}-1)\langle \ln {{\alpha }_{md}}\rangle -{{v}_{md}}{{{\langle{\alpha }}}_{md}\rangle} \right]}}\  \\
&  +\sum\limits_{m=1}^{M}{\left[ ({{s}_{m}}-1)\langle \ln {{\varphi }_{m}}\rangle -{{t}_{m}}\langle {{\varphi }_{m}}\rangle  \right]}+\text{Con}\text{.},
\end{split}
\end{equation}
where ${{\mathcal R}_{m}}={{\left\langle \ln \frac{\Gamma (\sum\nolimits_{d=1}^{D+1}{{{\alpha }_{md}}})}{\prod\nolimits_{d=1}^{D+1}{\Gamma ({{\alpha }_{md}})}} \right\rangle }}$.

With the mathematical expression in~\eqref{eq:22-0}, an analytically tractable solution is not feasible, which is due to the fact that ${\mathcal R}_{m}$ cannot be explicitly calculated (although it can be simulated by some numerical sampling methods). In order to apply~\eqref{eq:21} to explicitly calculate the optimal posterior distributions and with the principles of the EVI framework, it is required to introduce an auxiliary function $\tilde{\mathcal{R}_m}$ such that ${\mathcal{R}_m}\ge\tilde{\mathcal{R}}_m$.
According to, we can select ${\tilde{\mathcal R}_{m}}$ as
\begin{equation}\eqs\label{eq:24}
\begin{split}
& \tilde{\mathcal{R}}_m\\
=&\ln \frac{\Gamma (\sum\nolimits_{d=1}^{D+1}{{\langle{{{\alpha }}}_{md}\rangle}})}{\prod\nolimits_{d=1}^{D+1}{\Gamma ({\langle{{{\alpha }}}_{md}\rangle})}}+\sum\limits_{d=1}^{D+1}{\left[ \Psi (\sum\limits_{k=1}^{D+1}{{\langle{{{\alpha }}}_{md}\rangle}})-\Psi ({\langle{{{\alpha }}}_{md}\rangle}) \right]} \\
&\times \left[ \langle \ln {{\alpha }_{md}}\rangle -\ln {\langle{{{\alpha }}}_{md}\rangle} \right]{\langle{{{\alpha }}}_{md}\rangle},
\end{split}
\end{equation}
where $\Psi (\cdot )$ is the digamma function defined as $\Psi (a)=\partial\ln\Gamma (a)/\partial a$.

Substituting~\eqref{eq:24} into~\eqref{eq:22-0}, a lower bound to $\langle \ln {p}(\mathcal{X},\Theta )\rangle$ can be obtained as
\begin{equation}\eqs\label{eq:22}
\begin{split}
& \langle \ln {\tilde{p}}(\mathcal{X},\Theta )\rangle\\
 =&\sum\limits_{n=1}^{N}{\sum\limits_{m=1}^{M}{\langle {{z}_{nm}}\rangle }}\left[ {{{\tilde{\mathcal{R}}}}_{m}}\text{+}\sum\limits_{d=1}^{D}{(\langle{{{\alpha }}}_{md}\rangle-1)\ln {{x}_{nd}}} \right. \\
& \left. -\sum\limits_{d=1}^{D+1}{\langle{{{{\alpha }}}_{md}}\rangle}(1+\sum\limits_{d=1}^{D}{{{x}_{nd}}})+\langle \ln {{\lambda }_{m}}\rangle +\sum\limits_{j=1}^{m-1}{\langle \ln (1-{{\lambda }_{j}})\rangle }\right] \\
&+\sum\limits_{m=1}^{M}{\left[ \langle \ln {{\varphi }_{m}}\rangle +(\langle {{\varphi }_{m}}\rangle -1)\langle \ln (1-{{\lambda }_{m}})\rangle  \right]}\  \\
&+\sum\limits_{m=1}^{M}{\sum\limits_{d=1}^{D+1}{\left[ ({{u}_{md}}-1)\langle \ln {{\alpha }_{md}}\rangle -{{v}_{md}}{{{\langle{\alpha }}}_{md}\rangle} \right]}}\  \\
&  +\sum\limits_{m=1}^{M}{\left[ ({{s}_{m}}-1)\langle \ln {{\varphi }_{m}}\rangle -{{t}_{m}}\langle {{\varphi }_{m}}\rangle  \right]}+\text{Con}\text{.}.
\end{split}
\end{equation}
With~\eqref{eq:21}, we can get analytically tractable solutions for optimally estimating the posterior distributions of $\mathcal Z$, $\vec{\lambda }$, $\vec{\varphi}$, and $\Lambda$. We now consider each of these in more detail:
$\emph{{1)}}$ \emph{The posterior distribution of} $q(\mathcal Z)$

As any term that is independent of $z_{nm}$ can be absorbed into the additive constant, we have
\begin{equation}\eqs\label{eq:25}
\begin{split}
&\ln{{q^*}}({{z}_{nm}})=\text{Con.}+{{z}_{nm}}\left[ {{\widetilde{\mathcal{R}}}_{m}}+\langle \ln {{\lambda }_{m}}\rangle +\sum\limits_{j=1}^{m-1}{\langle \ln (1-{{\lambda }_{j}})\rangle } \right. \\
& \ \ \ \ \ \ \ \ +\sum\limits_{d=1}^{D}{({\langle{{{\alpha }}}_{md}\rangle}-1)\ln {{x}_{nd}}}+\left. \sum\limits_{d=1}^{D+1}{\langle{{{{\alpha }}}_{md}\rangle}\ln }(1+\sum\limits_{d=1}^{D}{{{x}_{nd}}}) \right],
\end{split}
\end{equation}
which has same logarithmic form of the prior distribution (\emph{i.e.}, the categorial distribution). Therefore, we can write $\ln q^*(\mathcal Z)$ as
\begin{equation}\eqs\label{eq:26}
\ln q^*(\mathcal Z)=\sum\limits_{n=1}^{N}{\sum\limits_{m=1}^{M}{{{z}_{nm}}}}\ln {{\rho }_{nm}}+\text{Con.}
\end{equation}
with the definition that
\begin{equation}\eqs\label{eq:27}
\begin{split}
& \ln {{\rho }_{nm}}=\langle \ln {{\lambda }_{m}}\rangle +\sum\limits_{j=1}^{m-1}{\langle \ln (1-{{\lambda }_{j}})\rangle }+{{{\tilde{\mathcal{R}}}}_{m}} \\
& \ \ \ \ \ \ \ \text{+}\sum\limits_{d=1}^{D}{({\langle{{{\alpha }}}_{md}\rangle}-1)\ln {{x}_{nd}}-\sum\limits_{d=1}^{D+1}{\langle{{{{\alpha }}}_{md}\rangle}}(1+\sum\limits_{d=1}^{D}{{{x}_{nd}}})}.
\end{split}
\end{equation}
Recalling that ${{z}_{nm}}\in (0,1)$ and $\sum\nolimits_{m=1}^{M}{{{z}_{nm}}}=1$, we define
\begin{equation}\eqs\label{eq:30}
{{r}_{nm}}=\frac{{{\rho }_{nm}}}{\sum\nolimits_{m=1}^{M}{{{\rho }_{nm}}}}.
\end{equation}
Taking the exponential of both sides of \eqref{eq:26}, we have
\begin{equation}\eqs\label{eq:28}
q^*(\mathcal Z)=\prod\limits_{n=1}^{N}{\prod\limits_{m=1}^{M}{r_{nm}^{{{z}_{nm}}}}},
\end{equation}
which is the optimal posterior distribution of $\mathcal Z$.

The posterior mean $\langle {{z}_{nm}}\rangle $ can be calculated as $\langle {{z}_{nm}}\rangle ={{r}_{nm}}$. Actually, the quantities $\{{{r}_{nm}}\}$ are playing a similar role as the  responsibilities in the conventional EM~\cite{30} algorithm.

$\emph{{2)}}$ \emph{The posterior distribution of} $q(\vec \lambda)$

The optimal solution to the posterior distribution of $\lambda_{m}$ is given by
\begin{equation}\eqs\label{eq:31}
\begin{split}
& \ln {{q}}({{\lambda }_{m}})=\text{Con}\text{.+}\ln {{\lambda }_{m}}\sum\limits_{n=1}^{N}{\langle {{z}_{nm}}\rangle }\  \\
& \ \ \ \ \ \ \ \ \ \ \ +\ln (1-{{\lambda }_{m}})\left[ \sum\limits_{n=1}^{N}{\sum\limits_{j=m+1}^{M}{\langle {{z}_{nj}}\rangle +\langle {{\varphi}_{m}}\rangle -1}} \right],
\end{split}
\end{equation}
which has the logarithmic form of the beta prior distribution. Hence, the optimal posterior distribution is
\begin{equation}\eqs\label{eq:32}
q(\vec{\lambda })=\prod\limits_{m=1}^{M}{\text {Beta}({{\lambda }_{m}};g_{m}^{*},h_{m}^{*})},
\end{equation}
where the hyperparameters $s_{m}^{*}$ and $q_{m}^{*}$ are
\begin{equation}\eqs\label{eq:33}
g_{m}^{*}=1+\sum\limits_{n=1}^{N}{\langle {{z}_{nm}}\rangle },\;\;\;\; h_{m}^{*}=\langle{{\varphi }_{m}}\rangle+\sum\limits_{n=1}^{N}{\sum\limits_{j=m+1}^{M}{\langle {{z}_{nj}}\rangle }}.
\end{equation}

$\emph{{3)}}$ \emph{The posterior distribution of} $q(\vec \varphi)$

For variable $\varphi_{m}$, we have
\begin{equation}\eqs\label{eq:34}
\ln q^*({{\varphi }_{m}})=\text{Con.}+{{s}_{m}}\ln {{\varphi }_{m}}+\left[ \langle \ln (1-{{\lambda }_{m}})\rangle -{{t}_{m}} \right]{{\varphi }_{m}}.
\end{equation}
It can be observed that~\eqref{eq:34} has the logarithmic form of the gamma prior distribution. By taking the exponential of the both sides of~\eqref{eq:34}, we have
\begin{equation}\eqs\label{eq:35}
q^*(\vec{\varphi })=\prod\limits_{m=1}^{M}{\text{Gam}({{\varphi }_{m}};s_{m}^{*},t_{m}^{*})},
\end{equation}
where the optimal solutions to the hyperparamters $s_{m}^{*}$ and $t_{m}^{*}$ are
\begin{equation}\eqs\label{eq:36}
s_{m}^{*}=1+{{s}^0_{m}},\; \; \; \;  t_{m}^{*}={{t}^0_{m}}-\langle \ln (1-{{\lambda }_{m}})\rangle,
\end{equation}
where ${s}^0_{m}$ and ${t}^0_{m}$ denote the hyperparameters initialized in the prior distribution, respectively.

$\emph{{4)}}$ \emph{The posterior distribution of} $q(\Lambda)$

Similar to the above derivations, for the variable $\alpha_{md}$, $1\le d\le D+1$, the optimal approximation to the posterior distribution is
\begin{equation}\eqs\label{eq:37}
\begin{split}
& \ln q^*({{\alpha }_{md}})\\
=&\left\{ \langle {{z}_{nm}}\rangle \left[ \Psi (\sum\limits_{k=1}^{D}{\langle{{{{\alpha }}}_{mk}\rangle}})-\Psi ({\langle{{{\alpha }}}_{md}\rangle}) \right]+{{u}^0_{md}}-1 \right\}  \times \ln {{\alpha }_{md}}\\
& -\left\{ {{v}^0_{md}}-\sum\limits_{n=1}^{N}{\langle {{z}_{nm}}\rangle \left[ \ln {{x}_{nd}}-\ln(1+\sum\limits_{d=1}^{D}{{{x}_{nd}}}) \right]} \right\}\times {{\alpha }_{md}} \\
 & +\text{Con}\text{.}
\end{split}
\end{equation}
Since the posterior distribution of ${\alpha }_{md}$ has the logarithmic form of gamma distribution, we have
\begin{equation}\eqs\label{eq:39}
q^*(\Lambda)=\prod\limits_{m=1}^{M}{\prod\limits_{d=1}^{D+1}{\text{Gam}(\alpha_{md};u_{md}^{*},v _{md}^{*})}},
\end{equation}
where the optimal solutions to the hyperparameters $u_{md}^{*}$ and $v_{md}^{*}$ are given by
\begin{equation}\eqs\label{eq:40}
u_{md}^{*}={{u}^0_{md}}+\sum\limits_{n=1}^{N}{\langle {{z}_{nm}}\rangle \left[ \Psi (\sum\limits_{k=1}^{K+1}{{\langle{{{\alpha }}}_{mk}\rangle}})-\Psi ({\langle{{{\alpha }}}_{md}\rangle}) \right]}{\langle{{\alpha }}_{md}\rangle}
\end{equation}
and
\begin{equation}\eqs\label{eq:41}
v_{md}^{*}={{v}^0_{md}}-\sum\limits_{n=1}^{N}{\langle {{z}_{nm}}\rangle \left[ \ln {{x}_{nd}}-\ln (1+\sum\limits_{d=1}^{D}{{{x}_{nd}}}) \right]}.
\end{equation}
In the above equations, ${u}^0_{md}$ and ${v}^0_{md}$ are the hyperparameters in the prior distribution and we set $x_{n,D+1}=1$.
The following expectations are needed to calculate the aforementioned update equations:
\begin{equation}\eqs\label{eq:43}
\begin{split}
\langle \ln (1-{{\lambda }_{m}})\rangle =&\Psi (h_{m}^{*})-\Psi (g_{m}^{*}+h_{m}^{*}), \\
\langle \ln {{\lambda }_{m}}\rangle =&\Psi (g_{m}^{*})-\Psi (g_{m}^{*}+t_{m}^{*}), \\
 \langle \ln {{\alpha }_{md}}\rangle =&\Psi (u_{md}^{*})-\ln v_{md}^{*},\\
\langle {{\varphi }_{m}}\rangle =\frac{s_{m}^{*}}{t_{m}^{*}},&\ {\langle{{{\alpha }}}_{md}\rangle}=\frac{u_{md}^{*}}{v_{md}^{*}}.
\end{split}
\end{equation}

\begin{figure*}[!th]
\vspace{-5mm}
\psfrag{x}[][]{\sps ${\text{Iter.}}\ \sharp$}
\psfrag{y}[][]{\sps ${{\text{E}}_{q(\Theta)}}[\ln p(\mathcal X,\Theta)]-{{\text{E}}_{q(\Theta)}}[\ln q(\Theta)]$}
     \centering
          \subfigure[\Captionsize Model A]{\includegraphics[width=.32\textwidth]{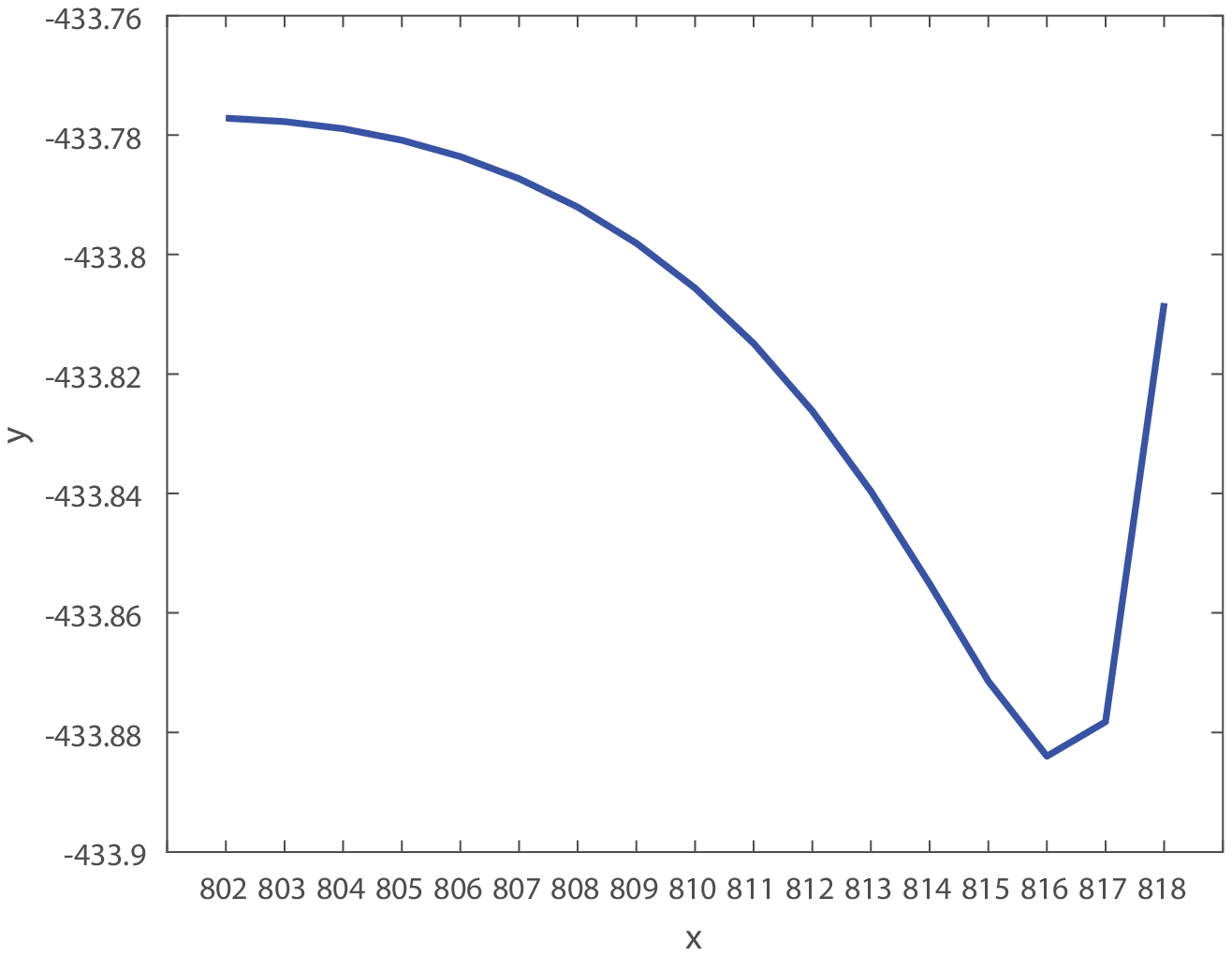}}
          \subfigure[\Captionsize Model B]{\includegraphics[width=.32\textwidth]{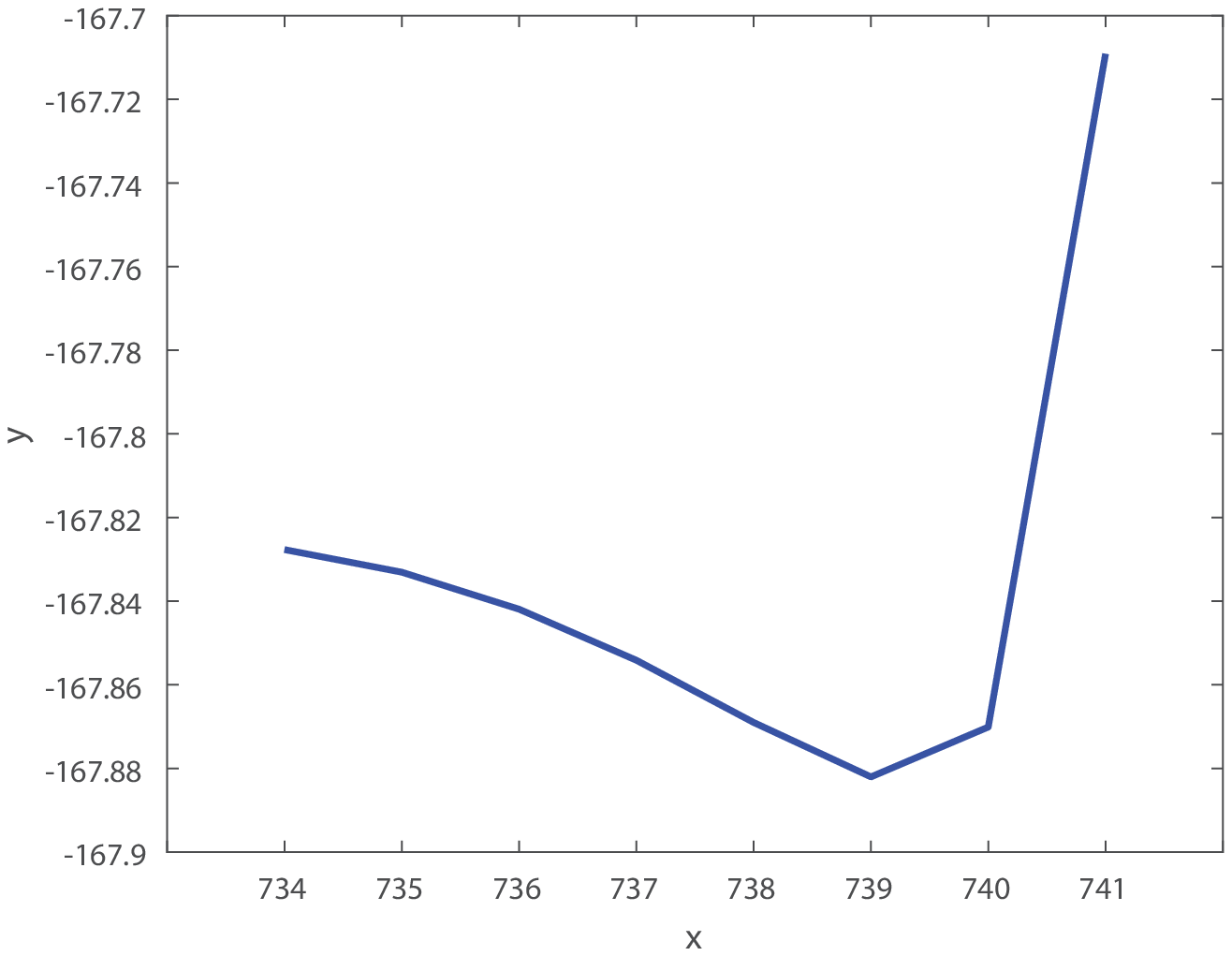}}
          \subfigure[\Captionsize Model C]{\includegraphics[width=.32\textwidth]{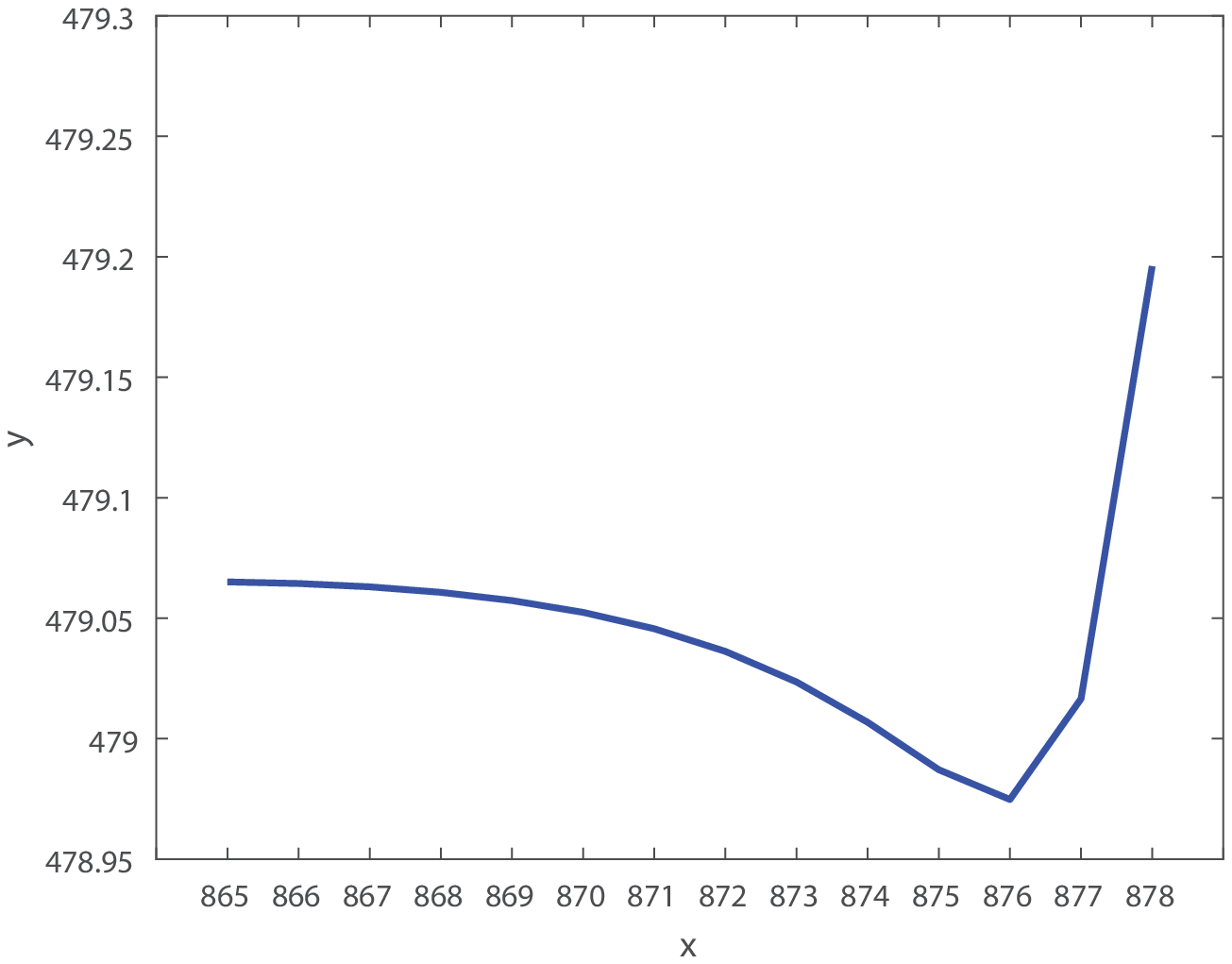}}
\vspace{-3mm}
          \caption{ \label{Fig: NonConvergence}\Captionsize Observations of the objective function's oscillations during iterations. This non-convergence indicates that the MLB approximation-based method cannot theoretically guarantee convergence.  The model settings are the same as Tab.~\ref{Tab: Model Estimation}. }
\vspace{-2mm}
\end{figure*}
\subsection{Full Variational Learning Algorithm}
As can be observed from the above updating process, the optimal solutions for the posterior distributions are dependent on the moments evaluated with respect to the posterior distributions of the other variables. Thus, the variational update equations are mutually coupled. In order to obtain optimal posterior distributions for all the variables, iterative updates are required until convergence. With the obtained posterior distributions, it is straightforward to calculate the lower bound $\tilde{\mathcal L}(q)$
\begin{equation}\eqs\label{eq:44}
\begin{split}
 \tilde{\mathcal{L}}(q)=&\int{q(\Theta )}\ln \frac{\tilde{p}(\Theta ,\mathcal X)}{q(\Theta )}d\Theta  \\
  =&\langle \ln \tilde{p}(\mathcal X,\Theta )\rangle -\langle \ln q(\Theta )\rangle  \\
  =&\langle \ln \tilde{p}(\mathcal{X},\Theta )\rangle -\langle \ln q(\mathcal{Z})\rangle -\langle \ln q(\vec{\lambda })\rangle  \\
&  -\langle \ln q(\vec{\varphi })\rangle -\langle \ln q(\Lambda)\rangle,
\end{split}
\end{equation}
which is helpful in monitoring the convergence. In~\eqref{eq:44}, each term with expectation (\emph{i.e.}, $\langle \cdot \rangle$) is evaluated with respect to all the variables in its argument as
\begin{equation}\eqs\label{eq:45}
\langle \ln q(\mathcal Z)\rangle ={{r}_{nm}}\ln {{r}_{nm}},
\end{equation}
\begin{equation}\eqs\label{eq:46}
\begin{split}
\langle \ln q(\vec{\lambda })\rangle =& \sum\limits_{m=1}^{M}{\left[ \ln \Gamma (g_{m}^{*}+h_{m}^{*})-\ln \Gamma (g_{m}^{*})-\ln \Gamma (h_{m}^{*}) \right.} \\
& \left. +(g_{m}^{*}-1)\langle \ln {{\lambda }_{m}}\rangle +(h_{m}^{*}-1)\langle \ln (1-{{\lambda }_{m}})\rangle  \right],
\end{split}
\end{equation}
\begin{equation}\eqs\label{eq:47}
\begin{split}
\langle \ln q(\vec{\varphi })\rangle =&\sum\limits_{m=1}^{M}{\left[ s_{m}^{*}\ln t_{m}^{*}-\ln \Gamma (s_{m}^{*}) \right.} \\
& \left. +(s_{m}^{*}-1)\langle \ln {{\varphi }_{m}}\rangle -t_{m}^{*}{{{\bar{\varphi }}}_{m}} \right],
\end{split}
\end{equation}
and
\begin{equation}\eqs\label{eq:48}
\begin{split}
\langle \ln q(\vec{\alpha })\rangle =&\sum\limits_{m=1}^{M}{\sum\limits_{d=1}^{D+1}{\left[ u_{md}^{*}\ln v_{md}^{*}-\ln \Gamma (u_{m}^{*}) \right.}} \\
& \left. +(u_{m}^{*}-1)\langle \ln {{\alpha }_{md}}\rangle -v_{md}^{*}{{{\bar{\alpha }}}_{md}} \right].
\end{split}
\end{equation}
Additionally, $\langle \ln \tilde{p}(\mathcal{X},\Theta )\rangle$ is given in~\eqref{eq:22} .

The algorithm of the proposed EVI-based Bayesian estimation of InIDMM is summarized in Algorithm~\ref{alg1}.
\begin{algorithm}[!t]
\caption{Algorithm for EVI-based Bayesian InIDMM}
 \label{alg1}
\begin{algorithmic}[1]
\STATE Set the initial truncation level ${M}$ and the initial values for hyperparameters $s^0_{m}$, $t^0_{m}$, ${{u}^0_{md}}$, and $v^0_{md}$ \\
\STATE Initialize the values of $r_{nm}$ by \emph{K}-means algorithm.\\
\REPEAT
\STATE  Calculate the expectations in $\eqref{eq:43}$.
\STATE  Update the posterior distributions for each variable by $\eqref{eq:33}$, $\eqref{eq:36}$, $\eqref{eq:40}$ and $\eqref{eq:41}$.
\UNTIL{Stop criterion is reached.}
\STATE For all $m$, calculate $\langle {{\lambda }_{m}}\rangle ={s_{m}^{*}}/{(s_{m}^{*}+t_{m}^{*})}\;$ and substitute it back into \eqref{eq: pi} to get the estimated values of the mixing coefficients $\widehat{\pi}_{m}$.
\STATE Determine the optimum number of components ${M}$ by eliminating the components with mixing weights smaller than ${{10}^{-5}}$.
\STATE Renormalize $\{\widehat{\pi}_{m}\}$ to have a unit $l_1$ norm.
\STATE Calculate $\widehat\alpha_{md} = u^*_{md}/v^*_{md}$ for all $m$ and $d$.
\end{algorithmic}
\end{algorithm}

\begin{table*}[!t]
\caption{Comparisons of true and estimated models.}
\label{Tab: Model Estimation}

\Tabsize
\centering
\begin{tabular}{c||c|c}
\hline
 True Model&\begin{tabular}{c}
 {Model A}\\
 $\pi_1=0.5\ \ ,\vec{\alpha}_1=[16\ 8\  6\ 2]^{\text{T}}$\\
 $\pi_2=0.5\ \ ,\vec{\alpha}_2=[8\  12\  15\  18]^{\text{T}}$
 \end{tabular}                  &\begin{tabular}{c}
 {Model B}\\
 $\pi_1=0.25\ \ ,\vec{\alpha}_1=[12\  36\  14\  18\  55\  16]^{\text{T}}$\\
 $\pi_2=0.25\ \ ,\vec{\alpha}_2=[32\  48\  25\   12\  36\  48]^{\text{T}}$\\
  $\pi_3=0.25\ \ ,\vec{\alpha}_3=[25\  10\  18\  10\  36\   48]^{\text{T}}$\\
 $\pi_4=0.25\ \ ,\vec{\alpha}_4=[6\  28\  16\   32\   12\  24]^{\text{T}}$
 \end{tabular} \\
 \hline
$\text{InIDMM}_{\text{SLB}}$ &\begin{tabular}{c}
 $\widehat{\pi}_1=0.502\ \ ,\widehat{\vec{\alpha}}_1=[16.96\ 8.58\ 6.39\ 12.49]^{\text{T}}$\\
 $\widehat{\pi}_2=0.498\ \ ,\widehat{\vec{\alpha}}_2=[8.20\ 12.16\ 15.49\ 18.34]^{\text{T}}$
 \end{tabular}   & \begin{tabular}{c}
 $\widehat{\pi}_1=0.251\ \ ,\widehat{\vec{\alpha}}_1=[12.26\ 36.59\ 14.30\ 18.19\ 56.36\ 16.25]^{\text{T}}$\\
 $\widehat{\pi}_2=0.249\ \ ,\widehat{\vec{\alpha}}_2=[33.37\ 49.92\ 25.85\ 12.80\ 37.00\ 49.79]^{\text{T}}$\\
  $\widehat{\pi}_3=0.252\ \ ,\widehat{\vec{\alpha}}_3=[25.72\ 10.32\ 18.09\ 10.09\ 37.27\ 49.58]^{\text{T}}$\\
 $\widehat{\pi}_4=0.248\ \ ,\widehat{\vec{\alpha}}_4=[6.14\ 28.94\ 16.72\ 33.46\ 12.32\ 25.20]^{\text{T}}$
 \end{tabular} \\
\hline
$\text{InIDMM}_{\text{MLB}}$ &\begin{tabular}{c}
 $\widehat{\pi}_1=0.508\ \ ,\widehat{\vec{\alpha}}_1=[15.20\ 7.71\ 5.90\ 11.64]^{\text{T}}$\\
 $\widehat{\pi}_2=0.492\ \ ,\widehat{\vec{\alpha}}_2=[9.21\ 13.76\ 17.13\ 21.10]^{\text{T}}$
 \end{tabular} & \begin{tabular}{c}
 $\widehat{\pi}_1=0.249\ \ ,\widehat{\vec{\alpha}}_1=[12.18\ 37.82\ 14.56\ 18.85\ 57.32\ 16.44]^{\text{T}}$\\
 $\widehat{\pi}_2=0.249\ \ ,\widehat{\vec{\alpha}}_2=[33.71\ 51.10\ 26.92\ 12.89\ 38.66\ 51.73]^{\text{T}}$\\
  $\widehat{\pi}_3=0.250\ \ ,\widehat{\vec{\alpha}}_3=[24.94\ 9.90\ 18.07\ 10.04\ 36.10\ 48.25]^{\text{T}}$\\
 $\widehat{\pi}_4=0.252\ \ ,\widehat{\vec{\alpha}}_4=[5.82\ 27.43\ 15.77\ 31.14\ 11.82\ 23.58]^{\text{T}}$
 \end{tabular} \\
\hline
\hline
True Model& \multicolumn{2}{|c}{\begin{tabular}{c}
 {Model C}\\
 $\pi_1=0.2\ \ ,\vec{\alpha}_1=[12\  21\  36\  18\ 32\  65\ 76]^\mathrm{T}$\\
 $\pi_2=0.2\ \ ,\vec{\alpha}_2=[28\  42\  21\  8\  54\   21\   48]^\mathrm{T}$\\
  $\pi_3=0.2\ \ ,\vec{\alpha}_3=[32\  12\  7\  35\  13\  32\  18]^\mathrm{T}$\\
 $\pi_4=0.2\ \ ,\vec{\alpha}_4=[62\  44\  31\  65\  72\   15\   44]^\mathrm{T}$\\
 $\pi_5=0.2\ \ ,\vec{\alpha}_5=[53\ 12\  18\  44\  65\  33\   52]^\mathrm{T}$
 \end{tabular}   }\\
 \hline
 $\text{InIDMM}_{\text{SLB}}$ &\multicolumn{2}{|c}{\begin{tabular}{c}
  $\widehat{\pi}_1=0.201\ \ , \widehat{\vec{\alpha}}_1=[12.08\ 20.89\ 36.25\ 18.28\ 32.69\ 65.72\ 76.70]^\mathrm{T}$\\
 $\widehat{\pi}_2=0.199\ \ ,\widehat{\vec{\alpha}}_2=[29.12\ 43.43\ 21.41\ 8.33\ 56.11\ 21.74\ 49.20]^\mathrm{T}$\\
  $\widehat{\pi}_3=0.200\ \ ,\widehat{\vec{\alpha}}_3=[31.57\ 11.89\ 6.99\ 34.70\ 12.90\ 31.85\ 17.89]^\mathrm{T}$\\
 $\widehat{\pi}_4=0.201\ \ ,\widehat{\vec{\alpha}}_4=[59.83\ 42.55\ 29.89\ 61.98\ 67.68\ 14.11\ 42.46]^\mathrm{T}$\\
 $\widehat{\pi}_5=0.199\ \ ,\widehat{\vec{\alpha}}_5=[58.00\ 12.8\ 20.02\ 47.70\ 71.08\ 36.57\ 57.66]^\mathrm{T}$
 \end{tabular}  }\\
 \hline
  $\text{InIDMM}_{\text{MLB}}$ &\multicolumn{2}{|c}{\begin{tabular}{c}
  $\widehat{\pi}_1=0.200\ \ , \widehat{\vec{\alpha}}_1=[12.56\ 21.50\ 37.69\ 19.00\ 33.06\ 68.04\ 79.64]^\mathrm{T}$\\
 $\widehat{\pi}_2=0.200\ \ ,\widehat{\vec{\alpha}}_2=[28.26\ 43.02\ 20.85\ 8.14\ 55.36\ 21.21\ 49.17]^\mathrm{T}$\\
  $\widehat{\pi}_3=0.199\ \ ,\widehat{\vec{\alpha}}_3=[32.17\ 12.19\ 7.13\ 35.66\ 13.01\ 32.54\ 17.84]^\mathrm{T}$\\
 $\widehat{\pi}_4=0.199\ \ ,\widehat{\vec{\alpha}}_4=[63.61\ 45.48\ 32.00\ 66.63\ 74.31\ 15.21\ 45.45]^\mathrm{T}$\\
 $\widehat{\pi}_5=0.202\ \ ,\widehat{\vec{\alpha}}_5=[52.12\ 11.83\ 18.34\ 43.77\ 64.80\ 32.53\ 51.48]^\mathrm{T}$
 \end{tabular}  }\\
 \hline
\end{tabular}
\vspace{-3mm}
\end{table*}
\section{Experimental Results}
\label{Sec: Experiments}
In this section, both synthesized data and real data are utilized to demonstrate the performance of the proposed algorithm for InIDMM. In the initialization stage of all the experiments, the truncation level $M$ is set to $15$ and the hyperparameters of the gamma prior distributions are chosen as ${{u}_{0}}={{s}_{0}}=1$ and ${{v}_{0}}={{t}_{0}}=0.005$, which provide non-informative prior distributions. Note that these specific choices were based on our experiments and were found convenient and effective in our case. We take the posterior means as point estimates to the parameters in an InIDMM.
\subsection{Synthesized Data Evaluation}
As shown in the previous studies for EVI-based Bayesian estimation, the single lower bound (SLB) approximation can guarantee the convergence while the multiple lower bound (MLB) approximation cannot. We use the synthesized data evaluation to compare the Bayesian InIDMM using the SLB approximation (proposed in this paper and denoted as $\text{InIDMM}_{\text{SLB}}$) with the Bayesian InIDMM using the MLB approximation (proposed in~\cite{Fan2015} and denoted as $\text{InIDMM}_{\text{MLB}}$). Three models (see Tab.~\ref{Tab: Model Estimation} for details) were selected to generate the synthesized datasets.

\begin{table*}[!t]
\vspace{-5mm}
\caption{Comparisons of objective function values and runtime for InIDMM with SLB and MLB.}
\label{Tab: SLBvsMLB4InIDMM}

\Tabsize
\centering
\begin{tabular}{@{}c@{}||cc|cc|cc}
\hline
\multirow{2}{*}{Model $\&$ Method}\ \ &\multicolumn{2}{c|}{Model A} &\multicolumn{2}{c|}{Model B} &\multicolumn{2}{c}{Model C}\\
\cline{2-7}
 &$\text{InIDMM}_{\text{SLB}}$ &$\text{InIDMM}_{\text{MLB}}$&$\text{InIDMM}_{\text{SLB}}$ &$\text{InIDMM}_{\text{MLB}}$&$\text{InIDMM}_{\text{SLB}}$ &$\text{InIDMM}_{\text{MLB}}$\\
\hline
\hline
Obj. Func. Val. &\ $\mathbf{-1.86\times 10^{3}}$\ \  & $-1.90\times 10^{3}$\ \ &\ $\mathbf{0.42\times 10^{3}}$\ \  & $0.32\times 10^{3}$\ \ &\ $\mathbf{3.05\times 10^{3}}$&\ \  $2.99\times 10^{3}$\\
\hline
$p$-values \ \ & \multicolumn{2}{c|}{$0.046$} &\multicolumn{2}{c|}{$6.48\times 10^{-4}$} &\multicolumn{2}{c}{$0.016$} \\
\hline
\hline
$\text{KL}(p(\mathcal{X}|\Theta)\|p(\mathcal{X}|\widehat{\Theta}))$ &\ $\mathbf{3.35\times 10^{-3}}$\ \  & $6.97\times 10^{-3}$\ \ &\ $\mathbf{2.80\times 10^{-3}}$\ \  & $8.07\times 10^{-3}$\ \ &\ $\mathbf{2.93\times 10^{-3}}$&\ \  $6.24\times 10^{-3}$\\
\hline
$p$-values \ \ & \multicolumn{2}{c|}{$1.46\times 10^{-11}$} &\multicolumn{2}{c|}{$6.93\times 10^{-15}$} &\multicolumn{2}{c}{$2.08\times 10^{-7}$} \\
\hline
\hline
Runtime (in $s$)$^\dagger$ &$\mathbf{2.06}$ & $2.26$  & $\mathbf{3.06}$ & $3.61$&$\mathbf{2.84}$ & $3.07$\\
\hline
\end{tabular}\\
$^\dagger$ \Captionsize On a ThinkCentre$^\circledR$ computer with Intel$^\circledR$ Core$^{\text{TM}}$ i$5-4590$ CPU $8$G.
\vspace{0mm}
\end{table*}

\begin{figure*}[!t]
\vspace{0mm}
\psfrag{x}[][]{\sps ${\text{Iter.}}\ \sharp$}
\psfrag{y}[][]{\sps ${{\text{E}}_{q(\Theta)}}[\ln p(\mathcal X,\Theta)]-{{\text{E}}_{q(\Theta)}}[\ln q(\Theta)]$}
\psfrag{c}[][]{\sps ${\text{Number of Mixture Component}}$}
     \centering
          \subfigure[\Captionsize Model A]{\includegraphics[width=.31\textwidth]{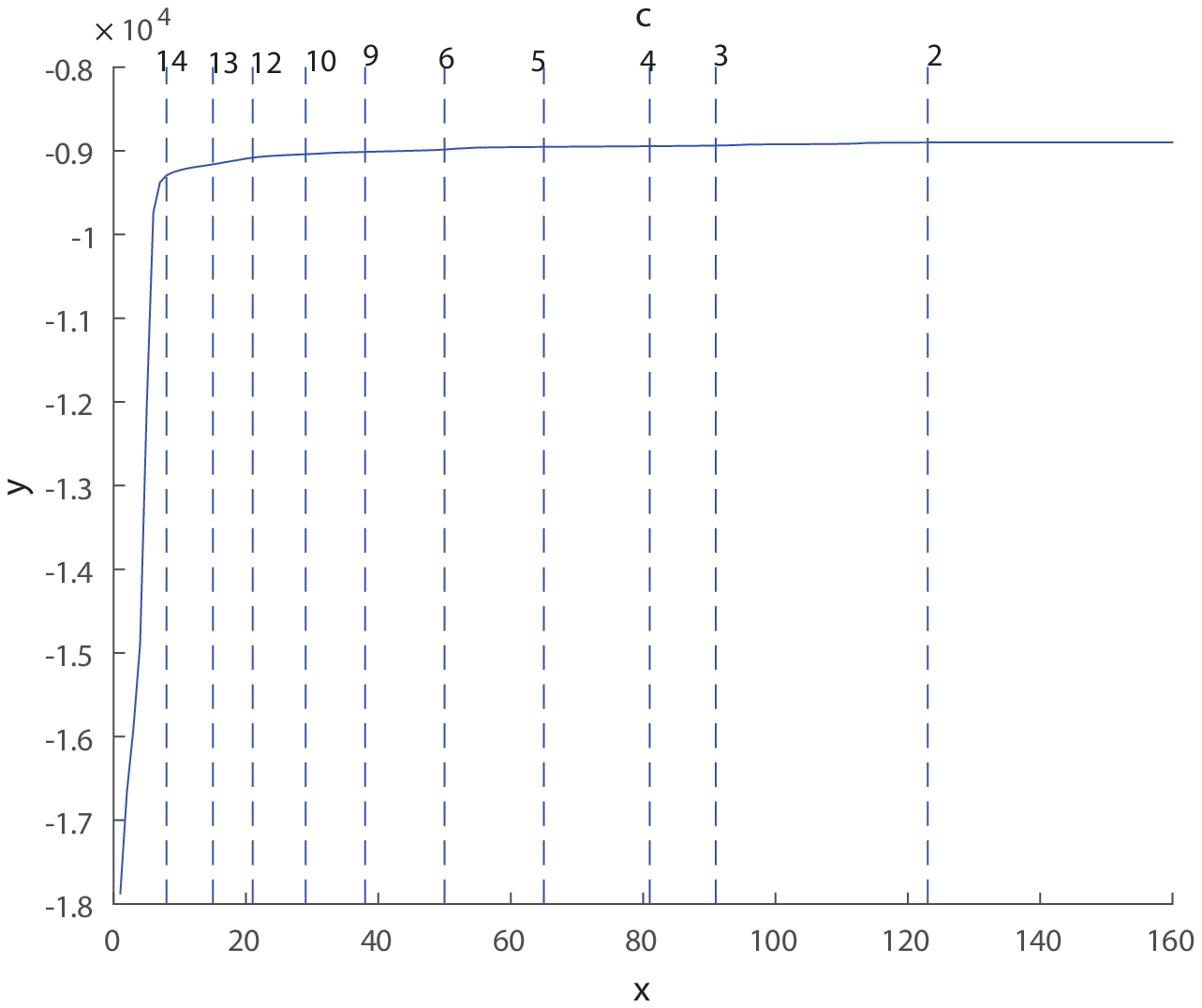}}\hspace{1mm}
          \subfigure[\Captionsize Model B]{\includegraphics[width=.32\textwidth]{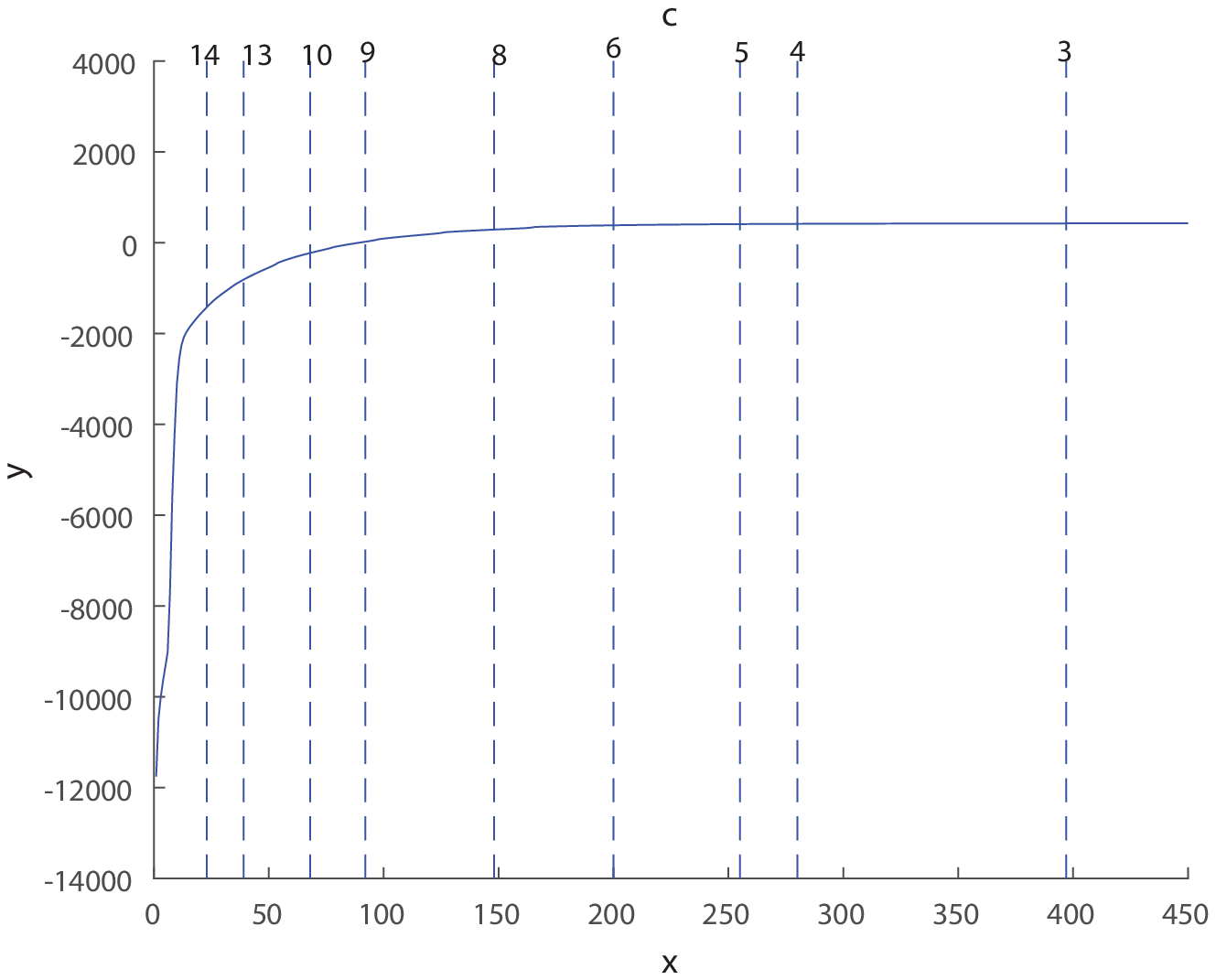}}\hspace{1mm}
          \subfigure[\Captionsize Model C]{\includegraphics[width=.32\textwidth]{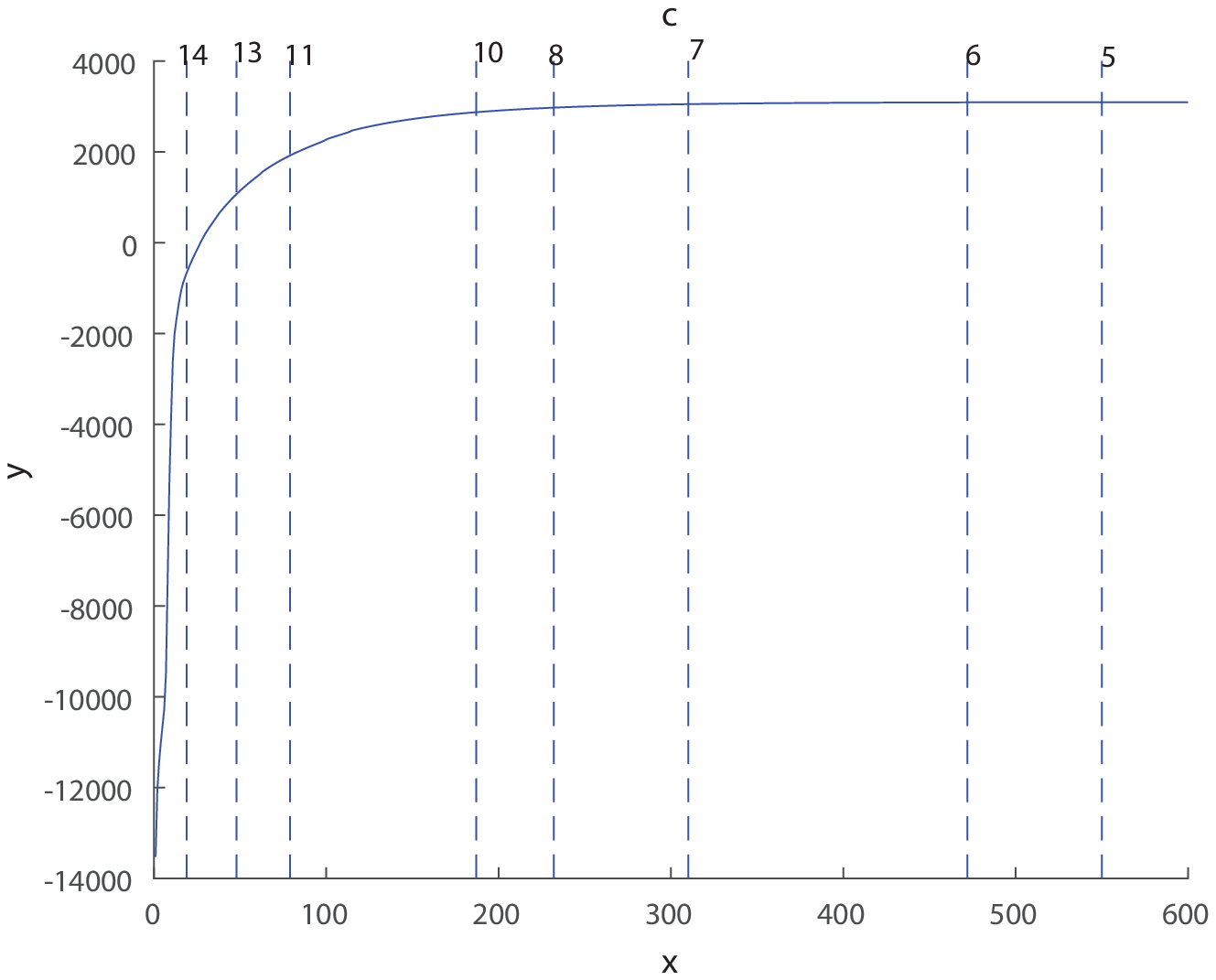}}
\vspace{-3mm}
          \caption{ \label{Fig: LBvsIter}\Captionsize Illustration of the variational objective function's values obtained by SLB against the number of iterations.}
\vspace{-1mm}
\end{figure*}

\begin{figure*}[!t]
\vspace{0mm}
\psfrag{a}[][]{\sps SLB}
\psfrag{b}[][]{\sps MLB}
\psfrag{c}[][]{\sps ${{\text{E}}_{q(\Theta)}}[\ln p(\mathcal X,\Theta)]-{{\text{E}}_{q(\Theta)}}[\ln q(\Theta)]$}
     \centering
          \subfigure[\Captionsize Model A]{\includegraphics[width=.32\textwidth]{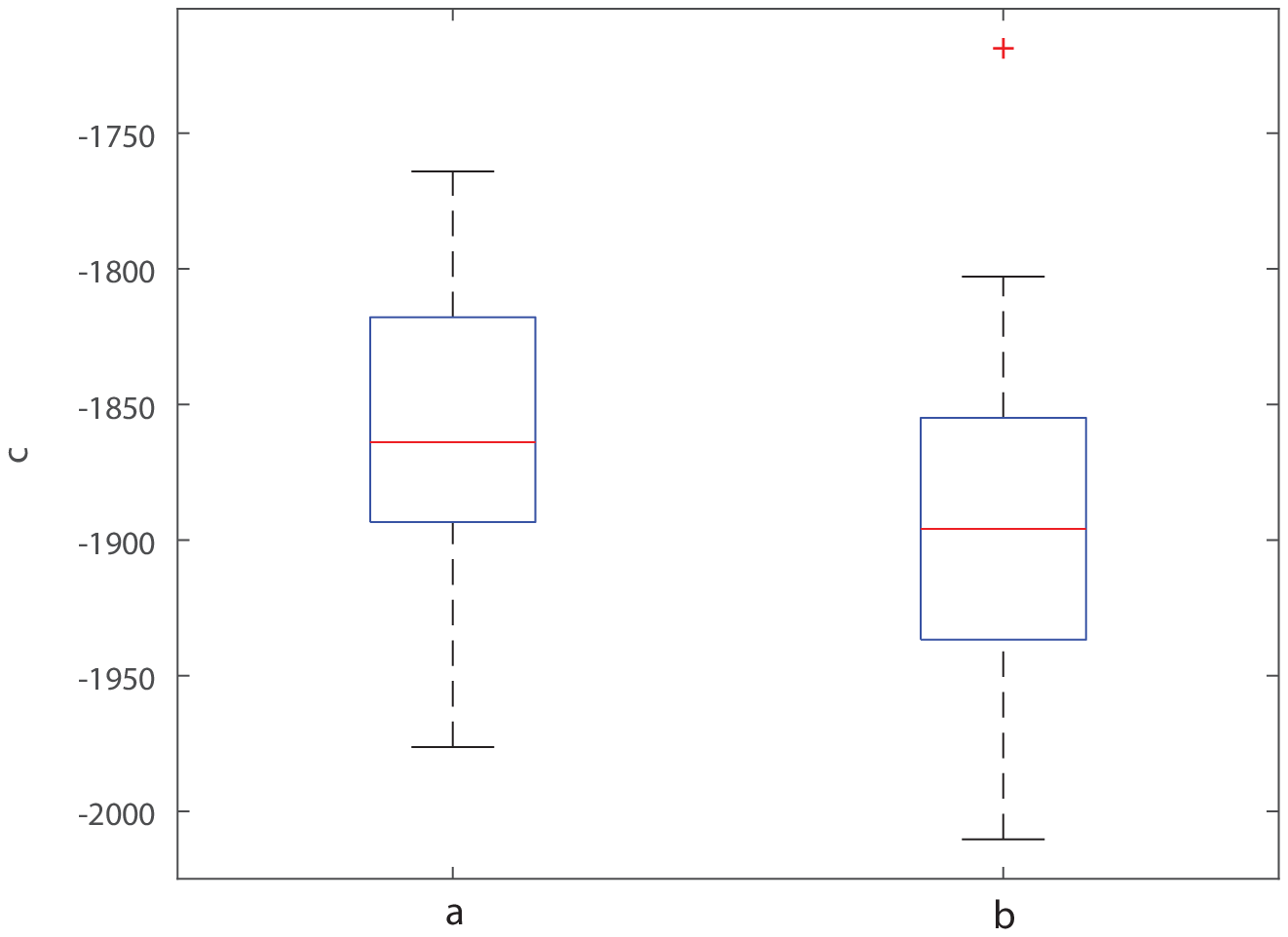}}
          \subfigure[\Captionsize Model B]{\includegraphics[width=.31\textwidth]{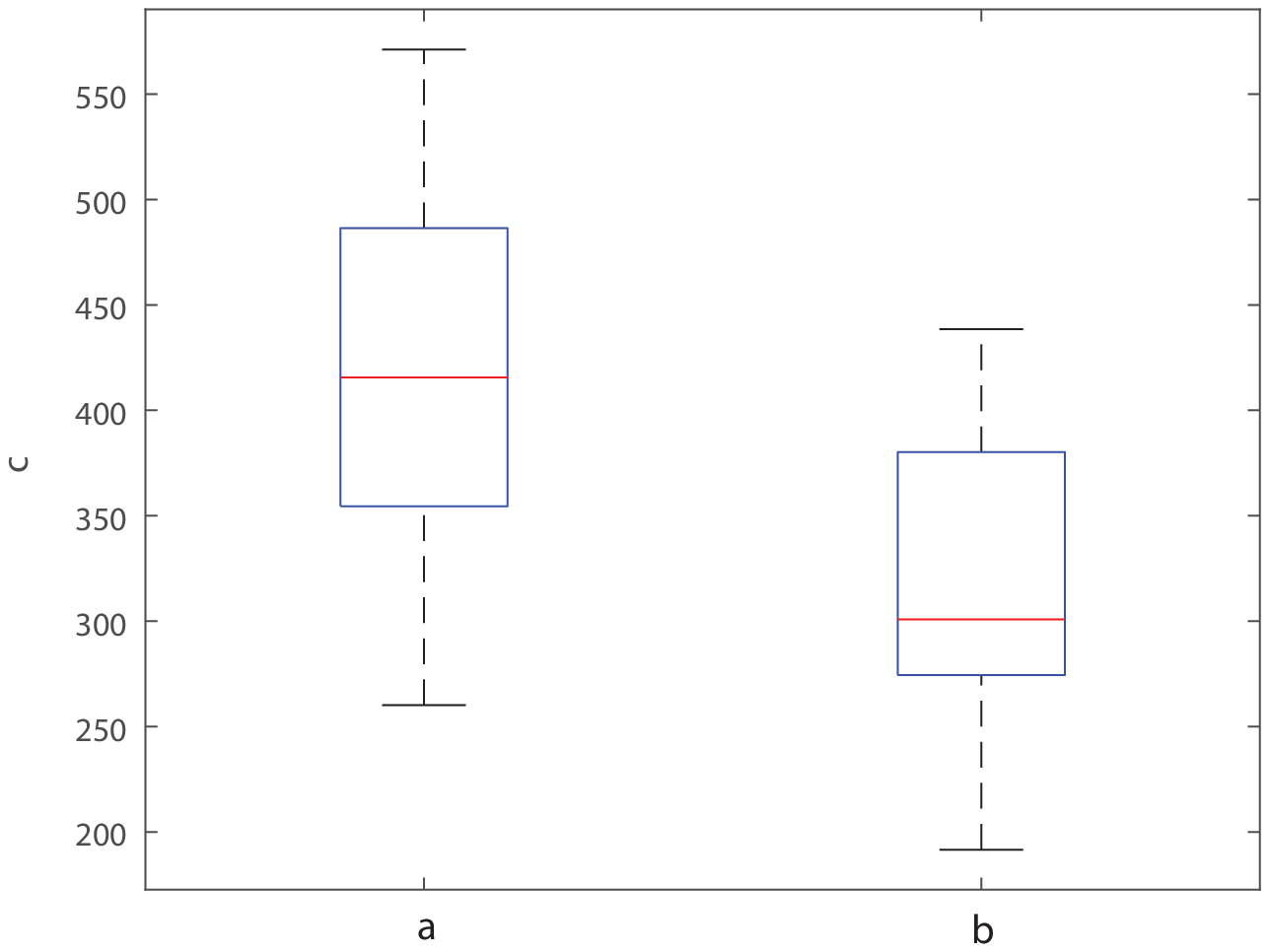}}
          \subfigure[\Captionsize Model C]{\includegraphics[width=.314\textwidth]{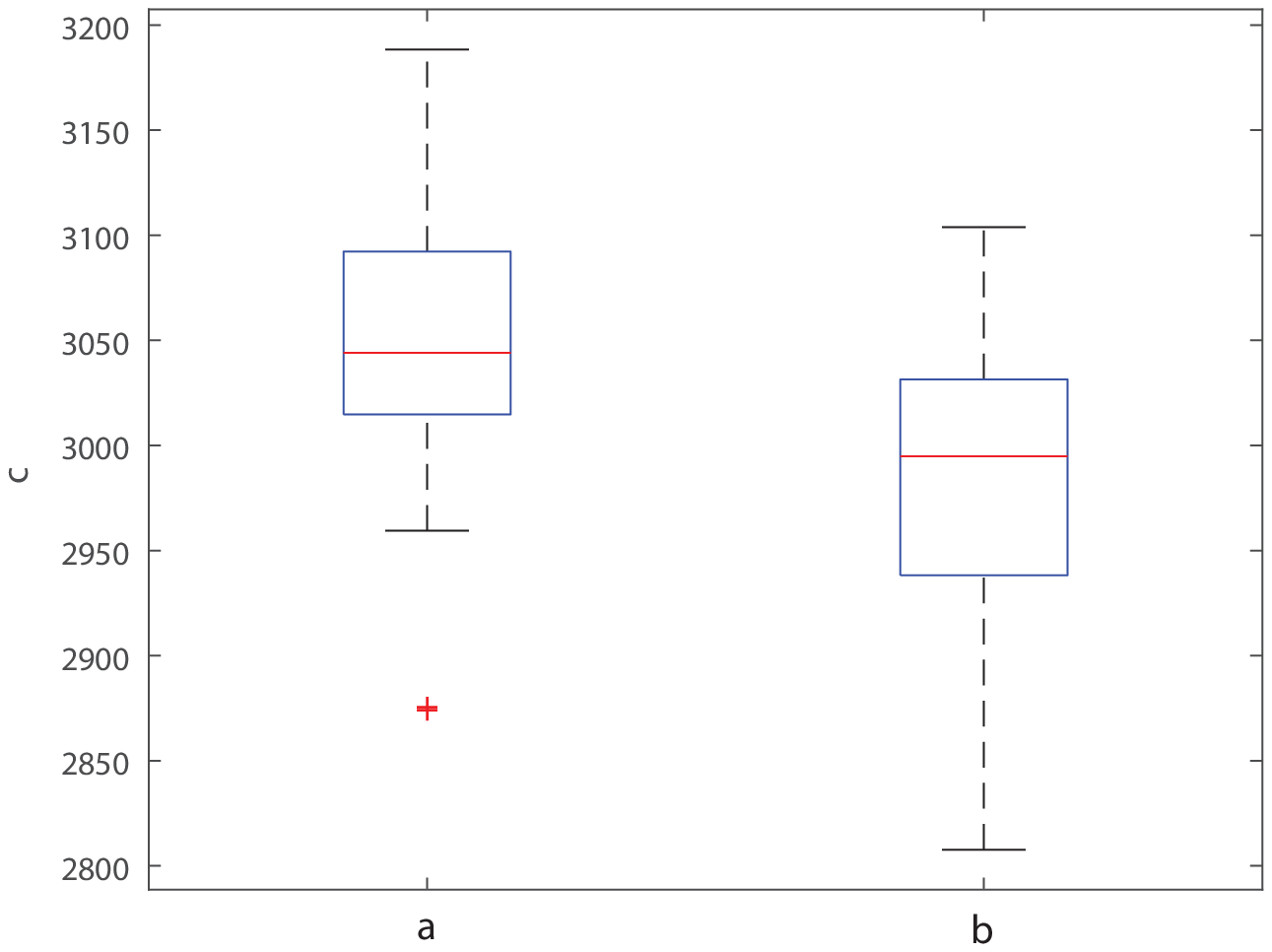}}
\vspace{-3mm}
          \caption{ \label{Fig: Boxplots4InIDMM}\Captionsize Boxplots for comparisons of the objective function values' distributions obtained by SLB and MLB with different models. The model settings are the same as those in Tab.~\ref{Tab: Model Estimation}. The central mark is the median, the edges of the box are the $25^{th}$ and $75^{th}$ percentiles. The outliers are marked individually.}
\vspace{-4mm}
\end{figure*}
\subsubsection{Observations of Oscillations}
We ran the $\text{InIDMM}_{\text{MLB}}$ algorithm and monitored the value of the variational objective function during each iteration. It can be observed that the variational objective function was not always increasing in Bayesian estimation with the $\text{InIDMM}_{\text{MLB}}$. Figure~\ref{Fig: NonConvergence} illustrates the decreasing values during iterations. On the other hand, the variational objective function obtained with the $\text{InIDMM}_{\text{SLB}}$ algorithm was always increasing until convergence, as the SLB approximation insures the convergency theoretically. The observations of oscillations demonstrate that the convergence with MLB approximation cannot be guaranteed. The original variational object function was numerically calculated by employing sampling method. In order to monitor the parameter estimation process of $\text{InIDMM}_{\text{SLB}}$, we show the value of the variational objective function during iterations in Fig.~\ref{Fig: LBvsIter}. It can be observe that the variational objective function obtained by $\text{InIDMM}_{\text{SLB}}$ increases during iterations and in most cases it increases very fast.

\subsubsection{Quantitative Comparisons}
Next, we compare the $\text{InIDMM}_{\text{SLB}}$ with the $\text{InIDMM}_{\text{MLB}}$ quantitatively. With a known IDMM, $2000$ samples were generated. The $\text{InIDMM}_{\text{SLB}}$ and the $\text{InIDMM}_{\text{MLB}}$ were applied to estimate the posterior distributions of the model, respectively. In Tab.~\ref{Tab: Model Estimation}, we list the estimated parameters  by taking the posterior means. It can be observed that, both the $\text{InIDMM}_{\text{SLB}}$ and the $\text{InIDMM}_{\text{MLB}}$ can carry out the estimation properly. However, with $20$ repeats of the aforementioned ``data generation-model estimation'' procedure and calculating the variational objective function with sampling method, superior performance of the $\text{InIDMM}_{\text{SLB}}$ over the $\text{InIDMM}_{\text{MLB}}$ can be observed from Tab.~\ref{Tab: SLBvsMLB4InIDMM}. The mean values of the objective function obtained by $\text{InIDMM}_{\text{SLB}}$ are larger than those obtained by the $\text{InIDMM}_{\text{SLB}}$ while the computational cost (measured in seconds) required by the $\text{InIDMM}_{\text{SLB}}$ are smaller than those required by the $\text{InIDMM}_{\text{MLB}}$. Moreover, smaller KL divergences\footnotemark\footnotetext{Here, the KL divergence is calculated as $\text{KL}(p(\mathcal{X}|\Theta)\|p(\mathcal{X}|\widehat{\Theta}))$ by sampling method. $\widehat{\Theta}$ denotes the point estimate of the parameters from the posterior distribution.} of the estimated models from the corresponding true models also verify that the $\text{InIDMM}_{\text{SLB}}$ yields better estimates than the $\text{InIDMM}_{\text{MLB}}$. In order to examine if the differences between the $\text{InIDMM}_{\text{SLB}}$ and the $\text{InIDMM}_{\text{MLB}}$ are statistically significant, we conducted the student's t-test with the null-hypothesis that the results obtained by these two methods have equal means and equal but unknown variances. All the $p$-values of in Tab.~\ref{Tab: SLBvsMLB4InIDMM} are smaller than the significant level $0.1$, which indicates that the superiority of the $\text{InIDMM}_{\text{SLB}}$ over the $\text{InIDMM}_{\text{MLB}}$ is statistically significant. The distributions of the objective function values are shown by the boxplots in Fig.~\ref{Fig: Boxplots4InIDMM}.

%
%

\begin{table}[!t]
\vspace{0mm}
\caption{\Captionsize Comparisons of image categorization accuracies (in $\%$) obtained with different models. The standard deviations are in the brackets. The $p$-values of the student's t-test with the null-hypothesis that $\text{InIDMM}_{\text{SLB}}$ and the referring method have equal means but unknown variances are listed.}
\label{Tab: Object categorization}
\Tabsize
\centering
\begin{tabular}{@{}c@{}||@{}c@{}c@{}c@{}c@{}}
\hline
 & $\text{InIDMM}_{\text{SLB}}$ & \ $\text{InIDMM}_{\text{MLB}}$ & $\text{IDMM}_{\text{SLB}}$ & $\text{IDMM}_{\text{MLB}}$\\
\hline
\hline
\ \ Caltech-$4$\ \ &  $\ \ \mathbf{93.49}(1.05)$\ \ & \ $92.27(1.91)$\ \ & \ $89.27(0.84)$\ \ & \ $88.75(2.04)$\ \  \\
$p$-value & N/A &\ $0.094$ &\ $1.01\times 10^{-8}$ &\ $3.79\times 10^{-6}$ \\
\hline
ETH-$80$ & $\ \mathbf{75.49}(0.75)$ &  $73.94(1.90)$ &  $72.88(1.46)$ &  $71.51(0.61)$\\
$p$-value & N/A &\  $0.027$&\ $8.69\times 10^{-5}$ &\ $1.25\times 10^{-10}$ \\
\hline
\end{tabular}
\vspace{-5mm}
\end{table}

\section{Conclusions}
\label{Sec: Conclusion}
The inverted Dirichlet distribution has been widely applied in modeling the positive vector (vector that contains only positive elements). The Dirichlet processing mixture of the inverted Dirichlet mixture model (InIDMM) can provide good modeling performance to the positive vectors. Compared to the conventional finite inverted Dirichlet mixture model (IDMM), the InIDMM has more flexible model complexity as the number of mixture components can be automatically determined. Moreover, the over-fitting and under-fitting problem is avoided by the Bayesian estimation of InIDMM. To obtain an analytically tractable solution for Bayesian estimation of InIDMM, we utilized the recently proposed extended variational inference (EVI) framework. With single lower bound (SLB) approximation, the convergence of the proposed analytically tractable solution is guaranteed, while the solution obtained via multiple lower bound (MLB) approximations may result in oscillations of the variational objective function. Extensive synthesized data evaluations and real data evaluations demonstrated the superior performance of the proposed method.

\bibliographystyle{IEEEtran}

\end{document}